\documentclass[runningheads]{llncs}

\usepackage{eccv}

\usepackage{eccvabbrv}

\usepackage{enumitem}
\usepackage{graphicx}
\usepackage{booktabs}
\usepackage{makecell}
\usepackage{arydshln} 
\usepackage{bbm}
\usepackage{bm}
\usepackage{subcaption} 
\usepackage{float}
\usepackage{caption}
\usepackage{afterpage}
\usepackage{algorithm}
\usepackage{algorithmic}
\usepackage[table]{xcolor}
\usepackage{pifont}
\usepackage{multirow}
\usepackage{amsmath} 
\usepackage{array}   
\usepackage{wrapfig}
\usepackage{xcolor}
\usepackage[accsupp]{axessibility}  

\usepackage[pagebackref,breaklinks,colorlinks,citecolor=eccvblue]{hyperref}

\usepackage{orcidlink}

\begin{document}

\title{Diversity-aware View Partitioning \\for Scalable VGGT} 

\titlerunning{Diversity-aware View Partitioning for Scalable VGGT}

\author{Jinsoo Park\inst{1} \quad
Donggyu Choi\inst{2} \quad
Ahyun Seo\inst{3} \quad
Minsu Cho\inst{1,4} \quad
Jeany Son\inst{1}\thanks{Corresponding author.}}
\authorrunning{J. Park \etal} 
\institute{%
$^1$POSTECH \quad $^2$GIST \quad $^3$KAIST \quad $^4$RLWRLD \\
\url{https://jspark1213.github.io/DA-VGGT}}
\maketitle

\begin{abstract}

Geometry transformers such as VGGT achieve strong performance by jointly reasoning over multiple views with global attention.
However, scaling them to large view collections remains challenging due to the quadratic cost of attention. 
Moreover, our empirical analysis reveals that the reconstruction quality in VGGT is sensitive to the distribution of viewpoints. 
Simply increasing the number of views without sufficient viewpoint diversity can even degrade performance, as redundant views introduce highly similar tokens that dilute informative geometric signals in the attention mechanism.
Motivated by this observation, we propose a training-free and plug-and-play VGGT inference framework that organizes views into diversity-aware balanced chunks. 
The chunks are constructed through combinatorial graph partitioning over visual dissimilarity and spatial dispersion. 
This view organization allows the transformer to focus attention on geometrically informative views while reducing redundant attention interactions.
To estimate spatial dispersion without full pose estimation, we approximate spatial relationships via a soft pose propagation strategy based on visual similarity from a small set of seed frames.
Extensive experiments demonstrate improved performance in camera pose estimation, multi-view depth prediction, and 3D reconstruction while reducing memory usage and inference latency. 
Our framework also complements existing VGGT variants, enabling scalable multi-view reconstruction without sacrificing geometric fidelity.

  \keywords{Multi-view Geometry \and Geometry Transformers \and Scalable Multi-view Reconstruction}
\end{abstract}
\section{Introduction}
\label{sec:intro}
Multi-view geometry transformers such as VGGT~\cite{VGGT} have recently emerged as a powerful paradigm for 3D scene reconstruction.
By jointly reasoning over multiple images through global attention, these models infer camera poses, multi-view depths, and pointmaps within a unified transformer architecture.
Despite this capability, VGGT remains fundamentally limited in scalability.
The core challenge arises from global attention: as the number of input frames grows, the token set grows proportionally, causing quadratic increases in computation and memory.
This quickly becomes prohibitive for long sequences, limiting the deployment of VGGT on thousands of images.
Recent work addresses this through various efficiency-oriented strategies~\cite{flashvggt,litevggt,vggtlong,swiftvggt,fast3r,hess}, reducing computation and memory overhead via token reduction~\cite{litevggt,fastvggt,tome,httm}, efficient attention~\cite{avggt,fastervggt,hess}, and hierarchical processing~\cite{mvp}.
However, these strategies alone remain insufficient for very long sequences.
Another line of work partitions long sequences into temporally ordered chunks processed sequentially~\cite{vggtlong,swiftvggt}, enabling reconstruction from larger collections while keeping memory manageable.
However, these pipelines rely on temporally ordered inputs and loop-closure constraints for global consistency, limiting their applicability when views are unordered.

\begin{figure}[t]
\centering
\includegraphics[width=1\linewidth]{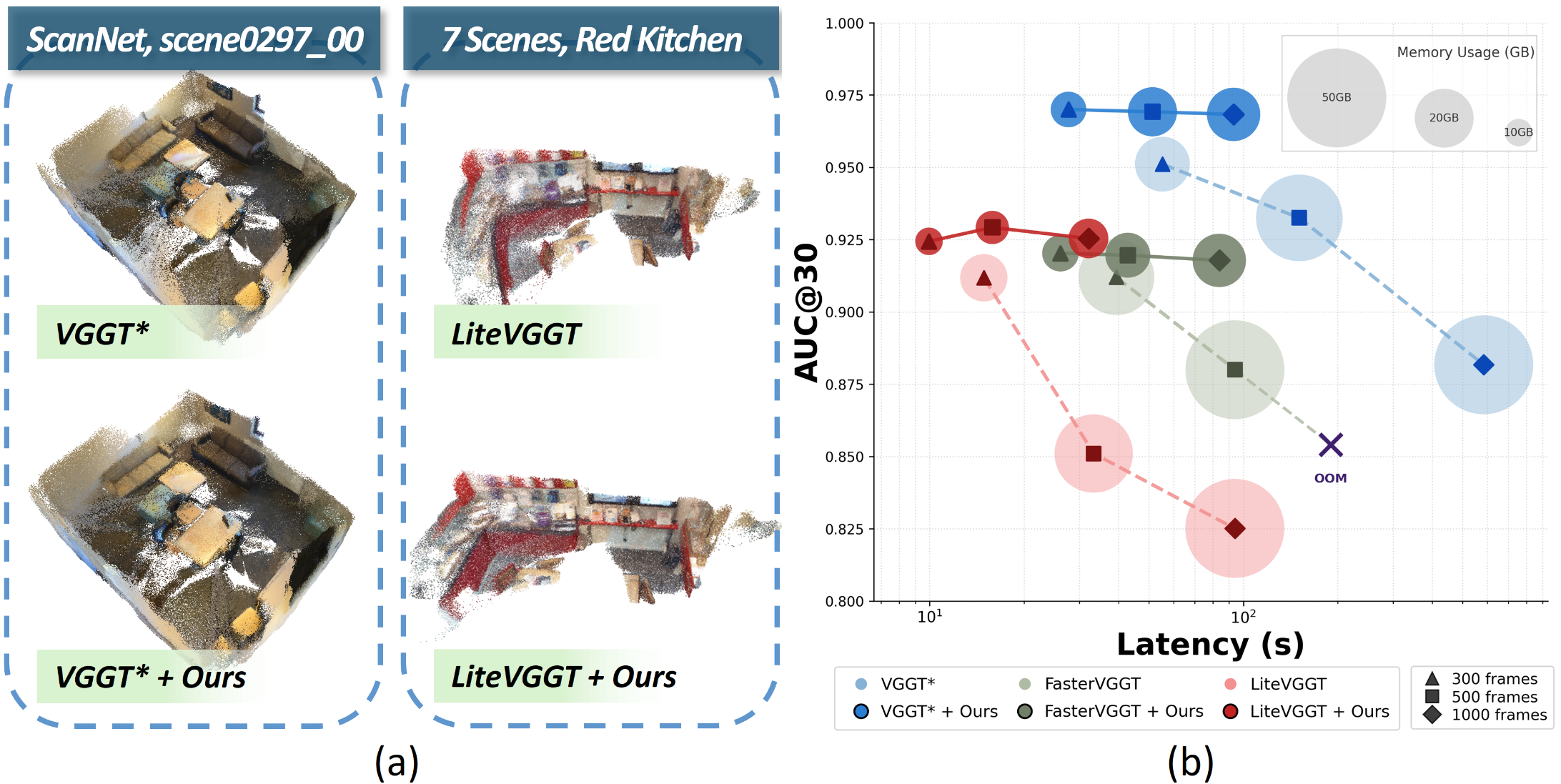}
\caption{ Effectiveness of diversity-aware view partitioning, 
(a) Our method reconstructs more geometrically consistent structures while preserving finer scene details compared to existing efficient VGGT variants, such as LiteVGGT~\cite{litevggt}.
(b) Unlike other methods~\cite{VGGT, litevggt}, whose performance tends to degrade as the number of input frames increases, our simple diversity-aware view partitioning mitigates this effect and achieves a better efficiency–accuracy trade-off.
}\label{fig:effi_chart} 
\end{figure}

Beyond these computational bottlenecks, we identify a more fundamental empirical phenomenon in VGGT: increasing the number of input views can even degrade reconstruction quality (Fig.~\ref{fig:effi_chart}).
Our observations suggest that this behavior stems from viewpoint redundancy and uneven coverage within the input view set.
In classical SfM, it is well known that near-duplicate views lead to small triangulation baselines and numerical instability. 
However, this issue manifests differently in VGGT, which performs multi-view reasoning through attention rather than explicit geometric triangulation.
In VGGT, redundant views reduce geometric diversity and dilute attention by introducing many non-informative tokens that overwhelm critical geometric cues such as parallax and occlusion, required for reliable pose and depth estimation.
These findings suggest that the key challenge in scaling VGGT lies not merely in the number of input views, but in the distribution of viewpoints provided to the model.

Motivated by this observation, we revisit VGGT from the perspective of view organization. Instead of asking how to process more views efficiently, we ask how views should be organized to maximize geometric reasoning under limited attention capacity.
To this end, we propose a training-free, plug-and-play framework that improves both reconstruction quality and computational efficiency through diversity-driven graph partitioning.
Rather than discarding or merging redundant views, our method reorganizes the entire collection into a set of geometrically diverse partitions. 
By promoting viewpoint diversity within each subset, the model can focus on complementary views and extract stronger geometric cues under limited attention capacity.
This leads to improved accuracy and scalability even for long, unordered sequences.

To construct these chunks, we formulate view partitioning as a combinatorial graph optimization problem that maximizes visual dissimilarity and spatial dispersion. 
Because spatial dispersion requires camera poses that are unavailable at initialization, we introduce a soft pose propagation strategy that estimates poses for a small set of seed frames and propagates them to the remaining frames based on visual similarity. 
This enables an efficient approximation of spatial diversity without additional forward passes or ground-truth data.

Extensive experiments show that our method improves camera pose estimation, multi-view depth estimation, and 3D reconstruction while reducing GPU memory consumption and inference latency compared to vanilla VGGT\footnote{In our experiments, we use the memory-efficient VGGT implementation from~\cite{fastvggt}, denoted as VGGT$^*$.}.
Our method with vanilla VGGT also outperforms existing efficiency-oriented VGGT variants~\cite{litevggt,fastervggt} in overall geometric performance. 
Unlike prior approaches that trade accuracy for efficiency, our method improves performance while reducing computational cost.
When combined with these variants, it further improves efficiency while maintaining competitive accuracy. 
Moreover, performance remains stable as the number of input frames increases, enabling scalable multi-view reconstruction in scenarios that previously resulted in out-of-memory failures.

Our contributions are summarized as follows:

\begin{itemize}
\item We reveal that viewpoint distribution plays a critical role in multi-view geometry transformers, showing that simply increasing the number of views without sufficient viewpoint diversity can degrade reconstruction performance.
\item We propose a training-free, plug-and-play framework for VGGT that reorganizes input views into diversity-aware balanced partitions via combinatorial graph optimization over visual dissimilarity and spatial dispersion.
\item To enable spatially informed partitioning without requiring camera poses, we introduce a soft pose propagation strategy that approximates spatial relationships from a small set of seed frames.
\item Our method improves camera pose estimation, multi-view depth prediction, and 3D reconstruction accuracy while reducing memory usage and inference latency, and further complements existing VGGT variants to enhance scalability.
\end{itemize}

\section{Related Work}
\label{sec:related}

\paragraph{\textbf{\textup{Feed-Forward Multi-View Models.}}}
Classical structure-from-motion (SfM) pipelines~\cite{glomap, vggsfm, colmap} rely on feature matching~\cite{lightglue,superglue,edm,loftr,croco}, geometric verification, and bundle adjustment to recover camera poses and 3D structure from unordered image collections.
While robust, these pipelines are computationally expensive and difficult to parallelize.
Recent work has moved toward feed-forward approaches that directly regress 3D quantities from image sets using learned transformers.
DUSt3R~\cite{dust3r}, MASt3R~\cite{mast3r} and Splatt3R\cite{splatt3r} predict dense pointmaps from image pairs, requiring global alignment to fuse pairwise predictions into a coherent reconstruction.
VGGT~\cite{VGGT} extends this paradigm to the multi-view setting: given $N$ input frames, a single forward pass through an alternating self- and cross-attention transformer jointly predicts camera poses, dense depth, and 3D point maps for all frames simultaneously.
While this joint reasoning enables strong performance, it incurs quadratic memory and compute scaling in attention, limiting the number of frames that can be processed in a single batch.

\paragraph{\textbf{\textup{Efficient VGGT Variants.}}}
Several concurrent works address the scalability bottleneck of VGGT through architectural modifications.
LiteVGGT~\cite{litevggt} and FastVGGT~\cite{fastvggt} reduce token count via progressive token merging, lowering per-layer cost at the expense of spatial resolution.
FasterVGGT~\cite{fastervggt} and AVGGT~\cite{avggt} replace dense global attention with sparse or approximate attention mechanisms that attend to only a subset of the most relevant tokens.
VGGT-Long~\cite{vggtlong} takes a system-level approach, partitioning temporally ordered sequences into overlapping chunks and stitching them via sequential alignment with loop closure, targeting kilometer-scale outdoor reconstruction.
In contrast, our approach operates at the input partitioning level and does not assume temporal ordering among frames.

\paragraph{\textbf{\textup{View Selection and Frame Partitioning.}}}
The problem of selecting informative views has a long history in active vision and SfM~\cite{colmap, glomap}, typically formulated as next-best-view selection~\cite{nbv_survey,nbv,gennbv,popnbv} or keyframe sampling~\cite{orbkf,splatamkf,adaptivekf}.
In the context of feed-forward models, the question shifts from which frames to acquire to how to \emph{partition} a fixed set of frames into chunks that each support high-quality reconstruction.
Existing chunking strategies for VGGT~\cite{vggtlong,flashvggt,swiftvggt} use temporal stride or sequential windowing, which does not account for the visual content of the frames.
We instead formulate frame partitioning as a combinatorial optimization problem over a pairwise visual distance graph derived from frozen DINOv2~\cite{dino} features.
Our approach draws on the Kernighan--Lin (KL) local search~\cite{KL} for balanced graph partitioning, adapting the classical 2-opt swap procedure to maximize within-chunk viewpoint diversity while maintaining balanced coverage across chunks.
To our knowledge, this is the first work to treat frame-to-chunk assignment as an explicit optimization problem for feed-forward multi-view models.
\section{Motivation}
\label{sec:motivation}

In practical multi-view capture scenarios, frames are often densely sampled along camera trajectories, resulting in 
long sequences containing many visually similar views.
While such dense sampling increases the number of observations, it also introduces substantial redundancy that can hinder effective multi-view reasoning.
This is reflected in practice:
as shown in Fig.~\ref{fig:effi_chart}, increasing the number of input frames often leads to not only significantly higher latency but also degraded reconstruction accuracy for existing VGGT variants~\cite{VGGT,litevggt,fastervggt}.

\begin{figure}[t]
\centering
    \includegraphics[width=1\linewidth]{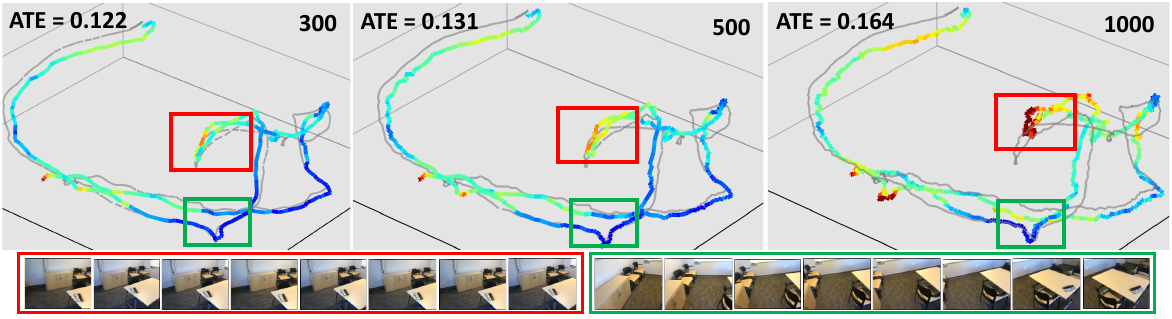}
    \caption{Impact of frame count on VGGT performance. Red boxes indicate high-error regions where VGGT struggles with many frames, while green
    boxes indicate low-error regions. Frames sampled from each region show that failures occur in visually redundant views, whereas successful regions exhibit greater viewpoint diversity.
}
\label{fig:motiv2}
\end{figure}

Motivated by this phenomenon, we analyze the behavior of VGGT under densely sampled sequences (Fig.~\ref{fig:motiv2}).
By examining sequences of the same scene captured at different frame rates, we observe that reconstruction errors tend to increase as the frame rate grows.
To investigate this,
we uniformly sample frames from regions with different reconstruction errors (red boxes indicate high-error regions, green boxes indicate low-error regions). 
We find that high-error regions often contain many visually redundant views, whereas low-error regions tend to exhibit more diverse viewpoints.

\begin{wrapfigure}[12]{r}{0.45\linewidth}
\vspace{-0.6cm}
\setlength{\intextsep}{0pt}
\centering
\includegraphics[width=\linewidth]{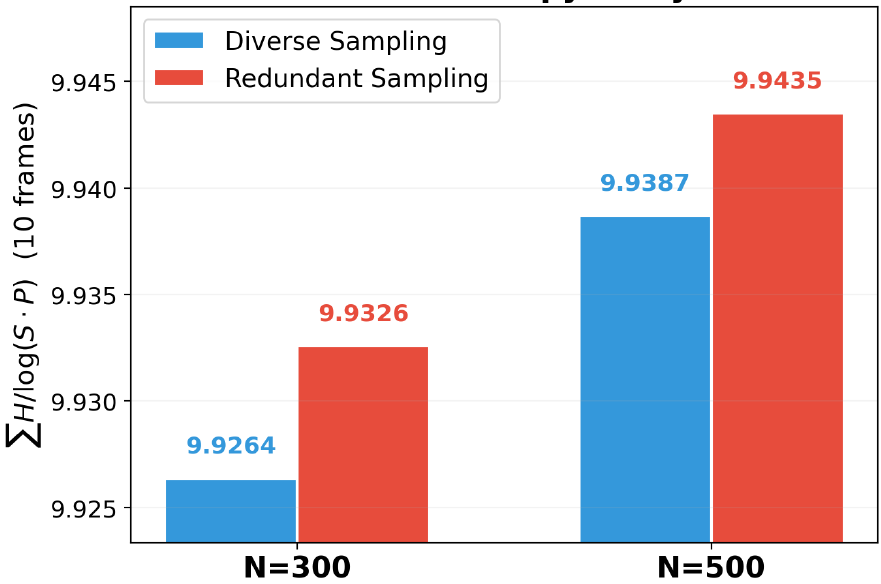}
\caption{Attention entropy analysis.}
\label{fig:attn_entropy}
\end{wrapfigure}
To better examine this behavior, we measure frame-level attention entropy under different frame counts and sampling strategies. 
From the same sequences, we construct subsets of 300 and 500 frames, each sampled either redundantly (visually similar frames) or sparsely (diverse viewpoints). 
As shown in Fig.~\ref{fig:attn_entropy}, the normalized attention entropy increases as the number of frames grows. 
Moreover, for the same frame count, redundantly sampled chunks exhibit higher entropy than sparsely sampled ones. 
This trend is consistent with attention dilution, where attention becomes more evenly distributed across many similar views rather than concentrating on geometrically informative ones.

These findings suggest that the main challenge in long sequences lies not in the number of frames but in how viewpoints are distributed: diversity is what matters most. Motivated by this insight, we reorganize the input views into balanced chunks that promote viewpoint diversity across frames,
enabling the transformer to focus on complementary views while mitigating attention dilution and improving both efficiency and reconstruction accuracy.
\section{Method}
\label{sec:method}

\begin{figure}[t]
\centering
\includegraphics[width=1\linewidth]{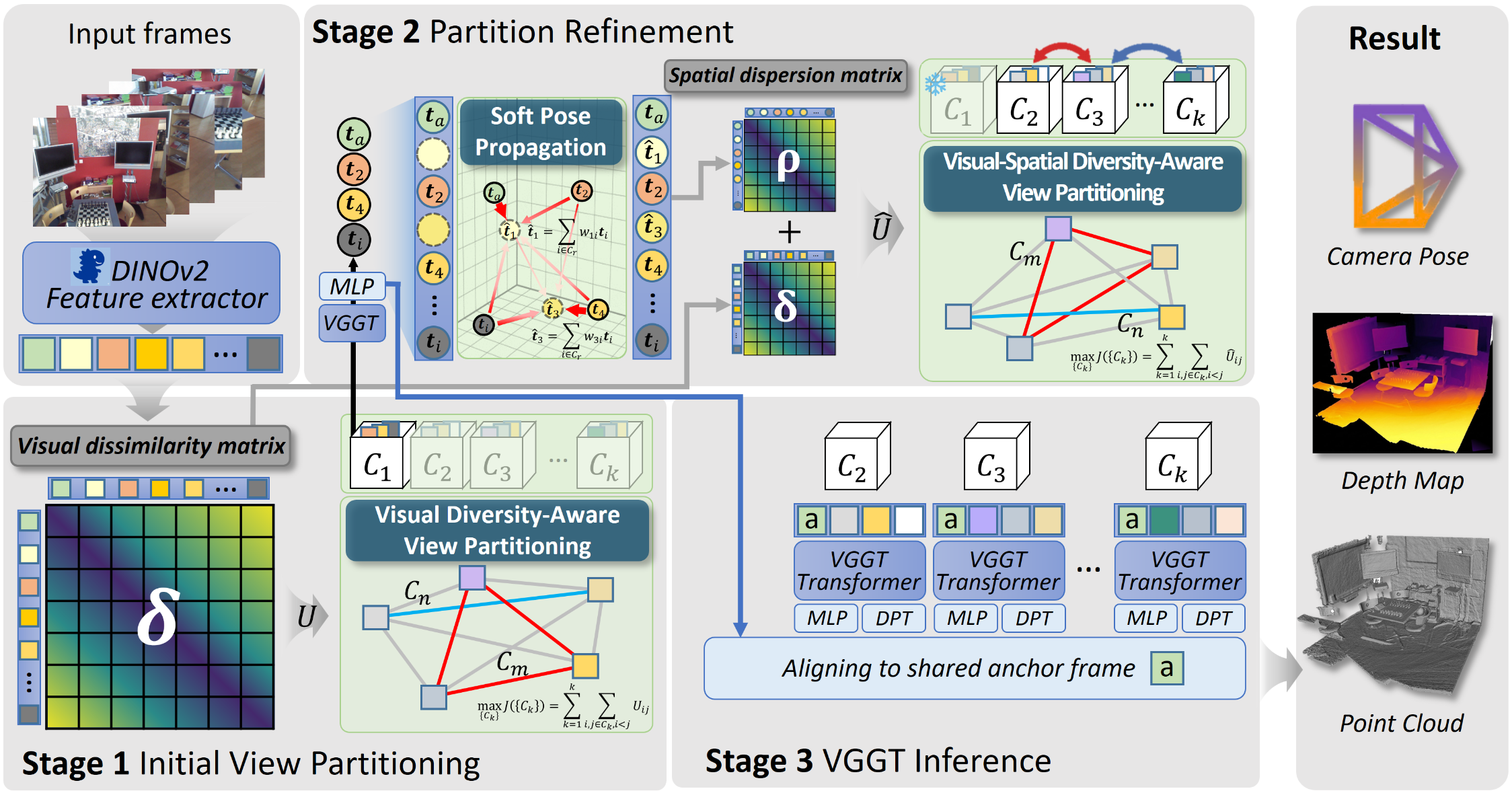}
\caption{Overview of the proposed framework.
Input frames are first partitioned using visual dissimilarity. A selected reference chunk is processed by VGGT to obtain pseudo-spatial cues, which are used to estimate pseudo-poses for the remaining frames. The frames are then reorganized into balanced chunks that maximize both visual and spatial diversity, and each chunk is processed independently by VGGT.} 
\label{fig:pipeline}
\end{figure}

Building on this insight, our goal is to reorganize the input views into balanced chunks that promoting viewpoint diversity, enabling efficient and reliable multi-view reasoning in geometry transformers. Rather than processing all views jointly, we partition them into chunks with complementary visual and spatial characteristics.

We formulate this as a diversity-aware balanced partitioning problem.
We first build an initial partition from visual dissimilarity, grouping $N$ frames into $K$ balanced chunks with high visual diversity.
Since camera poses are unavailable at initialization, we approximate spatial relationships via a lightweight soft pose propagation strategy based on visual similarity from a few seed frames.
The partition is then refined by a local search that jointly improves visual and spatial diversity while preserving balanced chunk sizes.
Finally, each chunk is processed independently by VGGT, and the predictions are aligned through a shared anchor frame for global consistency, as illustrated in Fig.~\ref{fig:pipeline}.

\subsection{VGGT Preliminaries}
\label{subsec:vggt_pre}

\paragraph{\textbf{\textup{Feature Backbone.}}}
Given a sequence of $N$ RGB images $\{I_i\}_{i=1}^{N}$, VGGT~\cite{VGGT} is a feed-forward transformer that jointly predicts per-frame 3D scene attributes in a single pass.
Each image is tokenized into patch embeddings by a frozen DINOv2~\cite{dino} encoder, then processed by $L{=}24$ layers of alternating frame-wise and global self-attention, followed by task-specific prediction heads. 
All outputs are expressed in the coordinate frame of the first (reference) view $I_1$.

\paragraph{\textbf{\textup{Prediction Heads.}}}
The \emph{camera head} takes the per-frame camera token and decodes it into a 9-dimensional pose encoding $\mathbf{g}_i \in \mathbb{R}^9$ comprising rotation and translation parameters, from which the full camera extrinsic $T_i \in \mathrm{SE}(3)$ and intrinsic $K_i \in \mathbb{R}^{3 \times 3}$ are recovered.
The \emph{depth head} and \emph{point head}, both based on DPT~\cite{dpt}, decode the patch tokens into dense depth maps in $\mathbb{R}^{H \times W}$ and point maps in $\mathbb{R}^{H \times W \times 3}$, respectively.

\subsection{Visual Diversity-Aware Initial View Partitioning}
\label{sec:kl_search}

Since camera poses are unavailable at initialization, this stage partitions frames using \emph{visual diversity} alone, deferring spatial information to Sec.~\ref{subsec:pseudo_pose}. We estimate pairwise visual dissimilarity between frames and use it to drive a balanced, diversity-maximizing partition.

\paragraph{\textbf{\textup{Visual Dissimilarity.}}}

To measure visual dissimilarity, we reuse the frozen feature extractor---the encoder built into VGGT~\cite{VGGT}---to obtain per-frame embeddings, and mean-pool the $M$ patch tokens of each frame into a single descriptor:
\begin{equation}
  \mathbf{z}_i = \frac{1}{M}\sum_{m=1}^{M} \mathbf{f}_{i,m}, \quad \mathbf{z}_i \in \mathbb{R}^d,
  \label{eq:meanpool}
\end{equation}
where $\mathbf{f}_{i,m}$ denotes the $m$-th patch token of frame $i$ produced by the feature extractor.

We compute the visual dissimilarity $\delta_{ij}$ between frames $i$ and $j$ as:
\begin{equation}
  \delta_{ij} = 1 - s_{ij}, ~~\text{where~~} s_{ij}=\frac{\mathbf{z}_i^\top \mathbf{z}_j}{\|\mathbf{z}_i\|\,\|\mathbf{z}_j\|}.
  \label{eq:vis_dis}
\end{equation}
Intuitively, frames with larger visual dissimilarity are more likely to correspond to different viewpoints or scene appearances, providing complementary information for multi-view reasoning.

\paragraph{\textbf{\textup{Balanced View Partitioning via Local Swap Optimization.}}}
Given $N$ input frames, we partition them into $K$ balanced chunks so that each chunk contains views with diverse visual and spatial characteristics.

Let $\mathcal{C}_k$ denote chunk $k$, i.e., the set of frames assigned to it, where $k \in \{1,\dots,K\}$, subject to $|\mathcal{C}_k| \leq c$.
We denote by $U_{ij}$ the diversity score between frames $i$ and $j$.
We formulate view partitioning as maximizing the total intra-chunk diversity score $\mathcal{J}$:
\begin{gather}
  \max_{\{\mathcal{C}_k\}} \mathcal{J}(\{\mathcal{C}_k\}) = \sum_{k=1}^{K} \sum_{i,j \in \mathcal{C}_k,\; i < j} U_{ij}, \label{eq:objective} \\ 
  \text{s.t. } \{\mathcal{C}_k\}\leq c,~~ \mathcal{C}_k\cap \mathcal{C}_{k'}=\emptyset,~ k\neq k',~~ 
  \bigcup_{k=1}^K \mathcal{C}_k=\{1,\dots,N\}. \nonumber
\end{gather}
Here, a score matrix $\mathbf{U} \in \mathbb{R}^{N \times N}$ is initialized with visual dissimilarity ($\mathbf{U}=\boldsymbol{\delta}$).

We call a partition \emph{balanced} when intra-chunk diversity is comparable across all chunks, so that no chunk is left with disproportionately redundant views. While Eq.~(\ref{eq:objective}) maximizes intra-chunk diversity, it does not by itself equalize this diversity across chunks; rather than adding a penalty term, we restrict the feasible set to balanced partitions and enforce balance structurally through the optimizer. Alg.~\ref{alg:kl_local_search} starts from a split satisfying the capacity bound $|\mathcal{C}_k|\leq\lceil N/K\rceil\leq c$ (with $K=\lceil N/c\rceil$) and updates only through size-preserving swaps, so this bound holds throughout. Balance arises from the same swaps: since each accepted swap has positive gain, it tends to move frames from lower- to higher-diversity chunks, equalizing intra-chunk diversity as the optimization converges.

Since solving this combinatorial problem exactly is expensive, we adopt a greedy local swap strategy inspired by the Kernighan–Lin algorithm~\cite{KL}.
Starting from a random balanced partition, we iteratively swap frame pairs across chunks to increase $\mathcal{J}$ while maintaining balanced sizes.
For chunks $(\mathcal{C}_{k_1},\mathcal{C}_{k_2})$, the gain of swapping $i \in \mathcal{C}_{k_1}$ and $j \in \mathcal{C}_{k_2}$ is
\begin{equation}
  \Delta_{ij} = \bigl(u_{k_2}(i) - u_{k_1}(i)\bigr) + \bigl(u_{k_1}(j) - u_{k_2}(j)\bigr) - 2\,U_{ij},
  \label{eq:swap_gain}
\end{equation}
where $u_k(i) = \sum_{m \in \mathcal{C}_k} U_{im}$.
We repeatedly apply swaps with positive gain until convergence.
The full procedure is summarized in Alg.~\ref{alg:kl_local_search}.

\begin{algorithm}[t]
\caption{$K$-way Balanced Graph Partitioning via Local Swap}
\label{alg:kl_local_search}
\begin{algorithmic}[1]
\REQUIRE Diversity score matrix $\mathbf{U} \in \mathbb{R}^{N \times N}$, chunks $K$, capacity $c$, iterations $T$
\ENSURE Partition $\{\mathcal{C}_1,\dots,\mathcal{C}_K\}$
\STATE Initialize $\{\mathcal{C}_k\}_{k=1}^K$ with a random balanced split
\FOR{$t = 1,\dots,T$}
    \STATE $\text{improved} \gets \texttt{false}$
    \FOR{each chunk pair $(k_1,k_2)$  with $k_1 < k_2$}
        \STATE $(i^*,j^*) \gets \arg\max_{i \in \mathcal{C}_{k_1},\, j \in \mathcal{C}_{k_2}} \Delta_{ij}$ \COMMENT{Eq.~\eqref{eq:swap_gain}}
        \STATE \textbf{if} $\Delta_{i^*j^*} > 0$ \textbf{then} swap $i^* \leftrightarrow j^*$; $\text{improved} \gets \texttt{true}$; \textbf{end if}
    \ENDFOR
    \STATE \textbf{if} {$\lnot\,\text{improved}$} \textbf{then} \textbf{break}; \textbf{end if}
\ENDFOR
\RETURN $\{\mathcal{C}_1,\dots,\mathcal{C}_K\}$
\end{algorithmic}
\end{algorithm}

\subsection{Spatial Diversity-Aware Partition Refinement}
Once approximate poses become available, this stage refines the partition along the \emph{spatial diversity} axis.
Since running VGGT on every frame would be costly, we estimate poses for a single reference chunk and propagate them to the remaining frames via visual similarity, then use the resulting spatial dispersion to refine the visual-only partition.

\label{subsec:pseudo_pose}
\paragraph{\textbf{\textup{Seed Chunk Pose Estimation.}}}
To approximate spatial relationships between views, we first estimate camera poses from a small reference chunk of frames. 
Specifically, after the initial visual diversity-based partitioning,
we simply select first chunk $\mathcal{C}_1$ as reference chunk ($\mathcal{C}_r$) and perform a single forward pass of VGGT to estimate their camera poses.
Let $\mathbf{p}_i = (\mathbf{R}_i, \mathbf{t}_i)$ denote the predicted pose of frame $i$ in this reference chunk $\mathcal{C}_r$.
These estimated poses serve as geometric anchors for the remaining frames. 

\paragraph{\textbf{\textup{Soft Pose Propagation.}}}
While camera poses can be estimated reliably within the reference chunk, the remaining frames do not have direct pose estimates. 
To approximate their spatial relationships without additional forward passes, we propagate geometric information from the reference chunk to the remaining frames using visual similarity.
For a frame $l \notin \mathcal{C}_r$, we approximate its pose by propagating pose information from the reference chunk using visual similarity:

\begin{equation}
\hat{\mathbf{t}}_l
=
\sum_{i \in \mathcal{C}_r}
w_{li} \mathbf{t}_i,  ~~\text{where~~} w_{li}=\frac{\exp(s_{li}/\gamma)}
{\sum_{j \in \mathcal{C}_r} \exp(s_{lj}/\gamma)}.
\label{eq:softmax}
\end{equation}
Here $s_{li}$ denotes the cosine similarity between frames $l$ and $i$ in the feature embedding space (Eq.~\eqref{eq:vis_dis}), and $\gamma$ controls the sharpness of the soft assignment.
Although these estimates are not intended to be accurate camera poses, they provide sufficient geometric cues to approximate spatial dispersion across views. 
Given pseudo poses $\hat{\mathbf{t}}_i$ for all $N$ frames, we then compute a pairwise spatial dispersion matrix $\boldsymbol{\rho}$:

\begin{equation}
  \rho_{ij} = 1 - \exp\!\left(-\frac{\|\hat{\mathbf{t}}_i - \hat{\mathbf{t}}_j\|}{\tau}\right) \in [0,1],
  \label{eq:pose_weight}
\end{equation}
where $\tau$ is set to the median pairwise distance to ensure scale invariance. 
This term assigns larger scores to pairs of frames that are spatially distant.

\paragraph{\textbf{\textup{Visual–spatial Diversity-aware View Partitioning.}}}
We incorporate this spatial dispersion into the diversity score matrix as
\begin{equation}
  \hat{U}_{ij} = \delta_{ij} + \epsilon\cdot {\rho}_{ij},
  \label{eq:combined_score}
\end{equation}
where $\epsilon$ controls the relative contribution of visual and spatial diversity.
Using the updated score matrix $\hat{\mathbf{U}}$, Eq.~\eqref{eq:objective} and Alg.~\ref{alg:kl_local_search} are applied to further refine the remaining $K{-}1$ chunks while keeping the reference chunk fixed.
This refinement yields visual–spatial diversity-aware view partitions that jointly improve visual diversity and spatial dispersion with minimal additional overhead.

\subsection{VGGT Inference on Balanced View Partitions}
\label{subsec:all_inference}
Given the updated partition 
$\{\mathcal{C}_k\}_{k=2}^K/\mathcal{C}_r$, we perform VGGT inference independently on each chunk, where the anchor frame $a$ is placed at the first position to serve as the reference view.
After first partitioning, each chunk is augmented with a shared \emph{anchor frame} (frame 0 by default) inserted at position~0.
This ensures that every chunk contains a common reference, enabling post-hoc alignment without any optimization.
After obtaining per-chunk predictions, we align all chunks to the reference chunk $C_r$ using the predicted pose of the shared anchor frame. We treat each pose $\mathbf{p}$ as an element of $SE(3)$, where $\cdot$ denotes composition and $(\cdot)^{-1}$ the inverse transform:
\begin{equation}
T^{(k)}_{\text{align}} = \hat{\mathbf{p}}^{(r)}_a \cdot (\hat{\mathbf{p}}^{(k)}_a)^{-1} \in SE(3),
\label{eq:alignment}
\end{equation}
All poses predicted in chunk $k$ are then transformed as 
$\tilde{\mathbf{p}}^{(k)}_i = T^{(k)}_{\text{align}} \cdot \hat{\mathbf{p}}^{(k)}_i$.
Because the anchor frame is shared across all chunks, this alignment provides a consistent global coordinate frame while avoiding additional global optimization.

\section{Experiments}
\label{sec:exp}

\subsection{Implementation Details}
\label{subsec:implementation}

\paragraph{\textbf{\textup{Datasets.}}}

We evaluate across six datasets spanning four tasks.
We assess camera pose estimation on 7Scenes~\cite{7scenes} and NRGBD~\cite{nrgbd}, multi-view depth on Bonn~\cite{bonn}, and 3D reconstruction on 7Scenes, NRGBD, and ScanNet-50~\cite{scannet}.
We further evaluate long-sequence SLAM on TUM-RGBD~\cite{tum} and outdoor reconstruction on Tanks\&Temples (T\&T)~\cite{tanks}.

\paragraph{\textbf{\textup{Metrics.}}}
We report AUC of $\max(\mathrm{RRA}, \mathrm{RTA})$ at $(3^\circ, 15^\circ, 30^\circ)$ for camera pose estimation, Abs Rel and $\delta<1.25$ for multi-view depth, and Chamfer distance (CD), accuracy (Acc.), completeness (Comp.), and normal consistency (NC) for 3D reconstruction.
For long-sequence and outdoor evaluation, we additionally report Absolute Trajectory Error (ATE), F1 at threshold $\tau$, and throughput (FPS).

\begin{table}[t]
\caption{Camera pose estimation on 7Scenes~\cite{7scenes}.}
\centering
\scriptsize
\resizebox{\linewidth}{!}{%
\begin{tabular}{c l r ccccc}
\toprule
{Frames} & \multicolumn{2}{c}{{Model}} &
{AUC@3}$\uparrow$ & {AUC@15}$\uparrow$ & {AUC@30}$\uparrow$ &
{VRAM(GB)}$\downarrow$ & {Latency(s)}$\downarrow$ \\
\midrule

\multirow{6}{*}[-1.5ex]{300} & \quad VGGT$^*$\cite{VGGT}        &         & 0.2412 & 0.6980 & 0.8269 & 18.76 & 54.16 \\
                    & \quad VGGT$^*$\cite{VGGT}            & + Ours  & \textbf{0.2481} & \textbf{0.7020} & \textbf{0.8292} & \textbf{12.68} & \textbf{27.65} \\
\cmidrule(lr){2-8}
                    & \quad FasterVGGT\cite{fastervggt} &        & 0.1662 & 0.6504 & 0.7992 & 25.18 & 38.87 \\
                    & \quad FasterVGGT\cite{fastervggt} & + Ours & \textbf{0.1760} & \textbf{0.6559} & \textbf{0.8024} & \textbf{12.98} & \textbf{25.99} \\
\cmidrule(lr){2-8}
                    & \quad LiteVGGT\cite{litevggt}    &         & \textbf{0.2234} & \textbf{0.6816} & \textbf{0.8165} & 16.43 & 15.35 \\
                    & \quad LiteVGGT\cite{litevggt}    & + Ours  & 0.2068 & 0.6698 & 0.8086 & \textbf{10.04} & \textbf{9.84} \\
\midrule

\multirow{6}{*}[-1.5ex]{500} & \quad VGGT$^*$\cite{VGGT}        &         & 0.2359 & 0.6942 & 0.8244 & 28.41 & 149.68 \\
                    & \quad VGGT$^*$\cite{VGGT}            & + Ours  & \textbf{0.2486} & \textbf{0.7013} & \textbf{0.8286} & \textbf{16.99} & \textbf{49.09} \\
\cmidrule(lr){2-8}
                    & \quad FasterVGGT\cite{fastervggt} &        & 0.1628 & 0.6462 & 0.7960 & 40.90 & 93.94 \\
                    & \quad FasterVGGT\cite{fastervggt} & + Ours & \textbf{0.1758} & \textbf{0.6546} & \textbf{0.8009} & \textbf{15.65} & \textbf{42.60} \\
\cmidrule(lr){2-8}
                    & \quad LiteVGGT\cite{litevggt}    &         & 0.2089 & 0.6691 & 0.8071 & 25.94 & 33.51 \\
                    & \quad LiteVGGT\cite{litevggt}    & + Ours  & \textbf{0.2117} & \textbf{0.6736} & \textbf{0.8108} & \textbf{11.91} & \textbf{15.97} \\
\midrule

\multirow{6}{*}[-1.5ex]{1000} & \quad VGGT$^*$\cite{VGGT}        &         & 0.2201 & 0.6853 & 0.8183 & 69.56 & 584.72 \\
                     & \quad VGGT$^*$\cite{VGGT}           & + Ours  & \textbf{0.2467} & \textbf{0.6998} & \textbf{0.8274} & \textbf{18.32} & \textbf{87.32} \\
\cmidrule(lr){2-8}
                     & \quad FasterVGGT\cite{fastervggt} &       & $OOM$ & $OOM$ & $OOM$ & $OOM$ & $OOM$ \\
                     & \quad FasterVGGT\cite{fastervggt} & + Ours & \textbf{0.1750} & \textbf{0.6534} & \textbf{0.7998} & \textbf{18.31} & \textbf{83.84} \\
\cmidrule(lr){2-8}
                     & \quad LiteVGGT\cite{litevggt}   &         & 0.1877 & 0.6198 & 0.7610 & 48.62 & 95.11 \\
                     & \quad LiteVGGT\cite{litevggt}   & + Ours  & \textbf{0.2083} & \textbf{0.6707} & \textbf{0.8087} & \textbf{13.98} & \textbf{31.68} \\
\bottomrule
\end{tabular}%
}
\label{tab:cam_7s}
\end{table}
\begin{table}[t]
\caption{Multi-view Depth Estimation on Bonn~\cite{bonn}.}
\centering
\scriptsize
\resizebox{0.8\linewidth}{!}{%
\begin{tabular}{c l r cccc}
\toprule
{Frames} & \multicolumn{2}{c}{{Model}} &
{Abs Rel}$\downarrow$ & {$\bm{\delta_1}$}$\uparrow$ &
{VRAM(GB)}$\downarrow$ & {Latency(s)}$\downarrow$ \\
\midrule
\multirow{6}{*}[-1.5ex]{500} & \quad VGGT$^*$\cite{VGGT}            &        & \textbf{0.048} & \textbf{0.961} & 28.53 & 156.47 \\
                    & \quad VGGT$^*$\cite{VGGT}                & + Ours & \textbf{0.048} & 0.954          & \textbf{15.78} & \textbf{65.84} \\
\cmidrule(lr){2-7}
                    & \quad FasterVGGT\cite{fastervggt}     &        & \textbf{0.046} & \textbf{0.959} & 41.02 & 102.6 \\
                    & \quad FasterVGGT\cite{fastervggt}     & + Ours & 0.047          & 0.956          & \textbf{15.78} & \textbf{49.92} \\
\cmidrule(lr){2-7}
                    & \quad LiteVGGT\cite{litevggt}         &        & 0.057          & \textbf{0.964} & 26.01 & 35.55 \\
                    & \quad LiteVGGT\cite{litevggt}         & + Ours & \textbf{0.055} & 0.956          & \textbf{11.98} & \textbf{14.69} \\
\bottomrule
\end{tabular}%
}
\label{tab:depth_bonn}
\end{table}

\paragraph{\textbf{\textup{Experimental Setup.}}}
We build on VGGT~\cite{VGGT} and run all experiments on a single NVIDIA A100 GPU (80 GB).
Following FastVGGT~\cite{fastvggt}, we retain activations only at layers $\{4, 11, 17, 23\}$ to reduce peak memory.
Since the original VGGT cannot exceed ${\sim}300$ frames on an 80 GB GPU, we use VGGT$^{*}$~\cite{fastvggt}, a memory-efficient re-implementation with comparable accuracy for longer inputs.
Operating only on input partitioning, our method applies without modification to LiteVGGT~\cite{litevggt} (token merging), FasterVGGT~\cite{fastervggt} (sparse attention), and the permutation-equivariant transformer $\pi^3$~\cite{pi3}, on which we further assess generalization against its efficient variants, FastVGGT~\cite{fastvggt} and FasterVGGT.

\subsection{Quantitative Results}

\paragraph{\textbf{\textup{Camera Pose Estimation.}}}
As shown in Tab.~\ref{tab:cam_7s}, our method reduces VRAM and latency across all VGGT variants while maintaining or improving pose accuracy.
On VGGT~\cite{VGGT}, it surpasses VGGT$^*$ at all sequence lengths, with up to $6.34\times$ speedup and $3.8\times$ lower VRAM, and higher AUC.
On FasterVGGT~\cite{fastervggt} and LiteVGGT~\cite{litevggt}, accuracy is preserved or improved on long sequences, with up to $3.5\times$ VRAM reduction and $3.0\times$ lower latency on long sequences.
Overall, our approach scales to long inputs: accuracy remains stable, while baselines degrade or run out of memory. At 1000 frames, FasterVGGT fails on an A100 (80\,GB), while our method enables all three variants to complete inference successfully.

\paragraph{\textbf{\textup{Multi-view Depth Estimation.}}}
Tab.~\ref{tab:depth_bonn} reports multi-view depth estimation on Bonn~\cite{bonn} at 500 frames.
Our method reduces VRAM by up to $2.6\times$ and latency by up to $2.4\times$ across all variants, while depth accuracy remains virtually unchanged---Abs Rel differences stay within 0.002 and $\delta_1$ within 0.008.
This confirms that the efficiency gains from our partitioning transfer to dense prediction tasks without compromising depth quality.

\begin{table}[t]
\caption{Multi-view 3D reconstruction on ScanNet-50~\cite{scannet}.}
\centering
\scriptsize
\resizebox{0.9\linewidth}{!}{%
\begin{tabular}{c l r ccccc}
\toprule
Frames & \multicolumn{2}{c}{Model} &
Acc.$\downarrow$ & Comp.$\downarrow$ & NC$\uparrow$ &
VRAM(GB)$\downarrow$ & Latency(s)$\downarrow$ \\
\midrule

\multirow{6}{*}[-1.5ex]{500} & \quad VGGT$^*$\cite{VGGT}             &        & 0.029 & 0.024 & \textbf{0.732} & 28.53 & 155.97 \\
                    & \quad VGGT$^*$\cite{VGGT}                 & + Ours & \textbf{0.025} & \textbf{0.019} & 0.724 & \textbf{19.13} & \textbf{59.73} \\
\cmidrule(lr){2-8}
                    & \quad FasterVGGT\cite{fastervggt}      &        & \textbf{0.032} & 0.024 & \textbf{0.703} & 41.02 & 99.87 \\
                    & \quad FasterVGGT\cite{fastervggt}      & + Ours & \textbf{0.032} & \textbf{0.021} & 0.682 & \textbf{15.78} & \textbf{51.94} \\
\cmidrule(lr){2-8}
                    & \quad LiteVGGT\cite{litevggt}          &        & 0.039 & 0.047 & \textbf{0.716} & 26.01 & 36.03 \\
                    & \quad LiteVGGT\cite{litevggt}          & + Ours & \textbf{0.029} & \textbf{0.018} & 0.700 & \textbf{11.98} & \textbf{17.28} \\
\bottomrule
\end{tabular}%
}
\label{tab:3d_scan}
\end{table}
\begin{table}[t]
\caption{Multi-view 3D reconstruction on 7Scenes~\cite{7scenes} and NRGBD~\cite{nrgbd}}
\centering
\scriptsize
\setlength{\tabcolsep}{5pt}
\resizebox{\linewidth}{!}{%
\begin{tabular}{c l r ccc ccc}
\toprule
\multirow{2}{*}{Frames} & \multicolumn{2}{c}{\multirow{2}{*}{Model}} &
\multicolumn{3}{c}{7Scenes} & \multicolumn{3}{c}{NRGBD} \\
\cmidrule(lr){4-6}\cmidrule(lr){7-9}
 & \multicolumn{2}{c}{} &
CD$\downarrow$ & NC$\uparrow$ & L(s)$\downarrow$ &
CD$\downarrow$ & NC$\uparrow$ & L(s)$\downarrow$ \\
\midrule

\multirow{6}{*}[-1.5ex]{500} & \quad VGGT$^*$\cite{VGGT}              &        & 0.045 & 0.648 & 155.38 & 0.037 & 0.810 & 155.95 \\
                    & \quad VGGT$^*$\cite{VGGT}                  & + Ours & \textbf{0.042} & \textbf{0.649} & \textbf{59.04} & \textbf{0.028} & \textbf{0.851} & \textbf{58.32} \\
\cmidrule(lr){2-9}
                    & \quad FasterVGGT\cite{fastervggt}       &        & 0.046 & \textbf{0.624} & 99.57 & 0.048 & 0.742 & 103.5 \\
                    & \quad FasterVGGT\cite{fastervggt}       & + Ours & \textbf{0.043} & 0.613 & \textbf{51.69} & \textbf{0.041} & \textbf{0.743} & \textbf{49.81} \\
\cmidrule(lr){2-9}
                    & \quad LiteVGGT\cite{litevggt}           &        & 0.049 & \textbf{0.626} & 34.48 & 0.069 & 0.743 & 40.18 \\
                    & \quad LiteVGGT\cite{litevggt}           & + Ours & \textbf{0.045} & 0.624 & \textbf{17.16} & \textbf{0.049} & \textbf{0.745} & \textbf{16.84} \\
\bottomrule
\end{tabular}%
}
\label{tab:3d_7s}
\end{table}

\paragraph{\textbf{\textup{Multi-view 3D Reconstruction.}}}
Tab.~\ref{tab:3d_scan} and Tab.~\ref{tab:3d_7s} evaluate 3D reconstruction on ScanNet-50~\cite{scannet}, 7Scenes~\cite{7scenes}, and NRGBD~\cite{nrgbd} at 500 frames.
On ScanNet-50, our method improves completeness for all baselines and maintains or improves accuracy with less VRAM and lower latency.
On VGGT~\cite{VGGT}, it improves Chamfer distance (CD) on both datasets while reducing latency by $2.6\times$.
For FasterVGGT~\cite{fastervggt} and LiteVGGT~\cite{litevggt}, it reaches competitive quality with up to $2.4\times$ speedups, sometimes surpassing the baseline in normal consistency (NC).
These results confirm that our diversity-aware chunking generalizes across tasks, improving dense 3D reconstruction alongside pose and depth estimation.


\begin{table}[t]
\caption{Pose estimation on TUM-RGBD~\cite{tum} and 3D reconstruction on T\&T~\cite{tanks}.}
\centering
\scriptsize
\setlength{\tabcolsep}{5pt}
\renewcommand{\arraystretch}{0.8}
\resizebox{\linewidth}{!}{%
\begin{tabular}{l l cccc}
\toprule
Dataset & Metric & Spann3R~\cite{spann3r} & MUSt3R-C~\cite{must3r} & VL~\cite{vggtlong} & VGGT~\cite{VGGT}+Ours \\
\midrule
\multirow{2}{*}{TUM-RGBD}
 & ATE [cm] $\downarrow$ & 38.74 & \underline{6.43} & 13.44 & \textbf{3.94} \\
 & FPS $\uparrow$        & \textbf{15.35} & \underline{10.25} & 1.32 & 6.32 \\
\cmidrule(lr){1-6}
\multirow{2}{*}{T\&T}
 & F1@$\tau\!\times\!10$ $\uparrow$ & 0.37 & 0.56 & \underline{0.63} & \textbf{0.74} \\
 & FPS $\uparrow$                   & \textbf{9.79} & 5.47 & 1.36 & \underline{7.61} \\
\bottomrule
\end{tabular}%
}
\label{tab:slam_mvs}
\end{table}

\begin{table}[t]
\caption{Generalization to $\pi^3$~\cite{pi3} on camera pose estimation on NRGBD~\cite{nrgbd}}
\centering
\scriptsize
\setlength{\tabcolsep}{5pt}
\resizebox{0.7\linewidth}{!}{%
\begin{tabular}{l ccc}
\toprule
Model & AUC@30$\uparrow$ & VRAM(GB)$\downarrow$ & L(s)$\downarrow$ \\
\midrule
$\pi^3$~\cite{pi3}                   & \textbf{0.9745}    & \underline{28.96} & 177.97            \\
\quad + FasterVGGT~\cite{fastervggt} & 0.9423             & 37.54             & 103.04            \\
\quad + FastVGGT~\cite{fastvggt}     & 0.9614             & 30.50             & \underline{72.56} \\
\quad + Ours                         & \underline{0.9717} & \textbf{7.95}     & \textbf{43.63}    \\
\bottomrule
\end{tabular}
}
\label{tab:pi3_eval}
\end{table}
\paragraph{\textbf{\textup{Comparison with Memory-based and Chunking Baselines.}}}
We evaluate long-sequence SLAM (TUM-RGBD~\cite{tum}) and outdoor reconstruction (Tanks\&Te\-mples~\cite{tanks}) against memory-based Spann3R~\cite{spann3r} and MUSt\-3R~\cite{must3r}, and chunking-based VGGT-Long (VL)~\cite{vggtlong}.
As shown in Tab.~\ref{tab:slam_mvs}, our method achieves the best accuracy on both benchmarks, reducing ATE by up to $3.4\times$ over VGGT-Long at competitive throughput, extending to outdoor and long-sequence regimes with a strong accuracy--efficiency trade-off.

\paragraph{\textbf{\textup{Generalization to Other Architectures.}}}
Since our framework operates only on input partitioning, it is not tied to VGGT~\cite{VGGT}.
Applying it to $\pi^3$~\cite{pi3} (Tab.~\ref{tab:pi3_eval}) preserves pose accuracy while cutting VRAM up to $3.6\times$ and latency up to $4.1\times$.
In contrast, methods that modify model internals, such as sparse attention (FasterVGGT~\cite{fastervggt}) and token merging (FastVGGT~\cite{fastvggt}), degrade $\pi^3$ accuracy with smaller gains, being tailored to a specific attention or tokenization scheme. This shows that organizing input views, rather than altering the network, generalizes beyond VGGT and its variants.
Note that we use FastVGGT~\cite{fastvggt} instead of LiteVGGT~\cite{litevggt} due to compatibility.

\subsection{Qualitative Results}
\setlength{\intextsep}{0pt}%
\begin{wrapfigure}[11]{r}{0.40\linewidth}
\centering
\includegraphics[width=\linewidth]{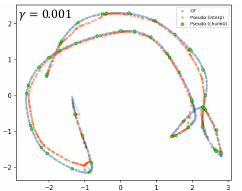}
\caption{Visualization of Pseudo-pose on NRGBD~\cite{nrgbd}.}
\label{fig:pos_tra}
\end{wrapfigure}
\paragraph{\textbf{\textup{Pseudo-Pose Estimation.}}}
Fig.~\ref{fig:pos_tra} shows the soft-assigned pseudo-positions: blue points are ground-truth camera centers, green points are VGGT predictions from the reference chunk, and red points are pseudo-positions for the remaining frames from Eq.~\eqref{eq:softmax}.
Our soft pose propagation closely follows the ground-truth trajectory, recovering the scene layout from a single chunk inference.
Although only approximate, these pseudo-positions provide useful geometric cues for partitioning when combined with appearance scores.

\begin{table}[t]
\centering
\renewcommand{\arraystretch}{1.1}
\begin{minipage}[t]{0.48\linewidth}
  \centering
  \caption{Partitioning strategy comparison on NRGBD~\cite{nrgbd}.}
  \label{tab:nrgbd_sampling}
  \scriptsize
  \setlength{\tabcolsep}{3pt}
  \begin{tabular}{l c c c}
  \toprule
  Methods & AUC@3$\uparrow$ & AUC@15$\uparrow$ & AUC@30$\uparrow$\\
  \midrule
  Random & 0.6819 & 0.8955 & 0.9412 \\
  Sequential & 0.3524 & 0.6559 & 0.7640 \\
  K-means & 0.3557 & 0.6644 & 0.7733\\
  Ours & \textbf{0.7844} & \textbf{0.9411} & \textbf{0.9691}\\
  \bottomrule
  \end{tabular}
\end{minipage}\hfill
\begin{minipage}[t]{0.48\linewidth}
  \centering
  \caption{Diversity-cue ablation on NRGBD~\cite{nrgbd}.}
  \label{tab:abl_ls}
  \scriptsize
  \setlength{\tabcolsep}{4pt}
  \begin{tabular}{l c c}
  \toprule
  Methods & AUC@3$\uparrow$ & {($+\Delta$)} \\
  \midrule
  Random Init. & 68.19 & \\
  w/ Spat. Disp. ($\boldsymbol{\rho}$) & 77.87 & {\textcolor[rgb]{0,0.55,0}{+ 9.68}} \\
  w/ Vis. dissim. ($\boldsymbol{\delta}$) & 78.21 & {\textcolor[rgb]{0,0.55,0}{+ 10.02}} \\
  w/ Both ($\boldsymbol{\delta}$+$\epsilon\boldsymbol{\rho}$) (Ours) & \textbf{78.44} & {\textcolor[rgb]{0,0.55,0}{+ \textbf{10.25}}} \\
  \bottomrule
  \end{tabular}
\end{minipage}
\end{table}

\subsection{Ablation Studies}
\paragraph{\textbf{\textup{Partitioning Strategy.}}}
Tab.~\ref{tab:nrgbd_sampling} compares different sequence partitioning strategies. Sequential partitioning collapses, as temporally adjacent frames are visually redundant and yield low-diversity chunks; K-means acts similarly, since it groups visually similar views together and thus \emph{minimizes} rather than maximizes within-chunk diversity. Random partitioning recovers much of the performance by breaking local redundancy, but still lacks explicit view organization. Our diversity-aware partitioning, which directly maximizes within-chunk dissimilarity, performs best across all AUC thresholds, confirming that view organization---not view count---is the source of the gains.

\paragraph{\textbf{\textup{Diversity cue.}}}
We adopt the KL algorithm~\cite{KL}, applying the best-improving swap at each step until no further improvement is found within a partition pair.
Despite this approximation, the objective $\mathcal{J}$ consistently converges within 5--10 outer iterations (See Sec.~\ref{subsec:supp_imple_mp} in supplementary).
The soft pose assignment of Eq.~(\ref{eq:softmax}) provides a further gain, as shown in Tab.~\ref{tab:abl_ls}, helping most in scenes where visually diverse viewpoints are spatially close, which appearance alone cannot distinguish.

\setlength{\intextsep}{6pt}
\begin{wraptable}{r}{0.46\linewidth}
\centering
\caption{Soft pose cue ablation on 7Scenes~\cite{7scenes}.}
\scriptsize
\setlength{\tabcolsep}{6pt}
\renewcommand{\arraystretch}{1.4}
\begin{tabular}{lccc}
\specialrule{0.8pt}{0pt}{0pt}
\rule{0pt}{2.4ex}Metric & T & T+R & R \\[0.4ex]
\hline
\rule{0pt}{2.6ex}AUC@3 & \textbf{0.2475} & 0.2361 & 0.2252 \\[0.4ex]
\specialrule{0.8pt}{0pt}{0pt}
\end{tabular}
\label{tab:poseweight_cue_s}
\end{wraptable}%
\paragraph{\textbf{\textup{Soft pose cue.}}}
We further analyze the pose cue in our soft pose assignment by comparing its rotation and translation terms, as shown in Tab.~\ref{tab:poseweight_cue_s}.
Since rotation captures viewing-direction similarity rather than camera-center dispersion, we ablate it to isolate the spatial signal: translation alone (T) outperforms both rotation alone (R) and their combination (T+R), confirming that camera-center translation is the most effective spatial cue for partitioning.

\paragraph{\textbf{\textup{Chunk Size Analysis.}}}
The chunk capacity $c$ controls the trade-off between reconstruction quality and computational cost, and—since our partitioning promotes within-chunk diversity—also governs the diversity–sparsity balance. Smaller chunks lower per-chunk complexity and improve scalability but sample each chunk more sparsely, increasing spatial sparsity; larger chunks provide denser coverage at higher memory and compute cost. Sparsity is thus not adjusted by the method itself but exposed through the chunk-size choice, while the balanced-partition constraint applies it uniformly across chunks. We evaluate this on 7Scenes~\cite{7scenes} in Tab.~\ref{tab:chunk_ablation} and further analyze the trade-off between chunk diversity and sparsity in the supplementary (Tab.~\ref{tab:sparsity}).
\begin{table}[t]
\caption{Chunk capacity ablation on (VGGT$^*$~\cite{VGGT} + Ours, 7Scenes~\cite{7scenes}).}
\centering
\scriptsize
\setlength{\tabcolsep}{6pt}
\begin{tabular}{c c ccccc}
\toprule
{Frames} & {Size} &
{AUC@3}$\uparrow$ & {AUC@15}$\uparrow$ & {AUC@30}$\uparrow$ &
{VRAM(GB)}$\downarrow$ & {Latency(s)}$\downarrow$ \\
\midrule
\multirow{3}{*}{1000}
 & 25  & 0.2399 & 0.6951 & 0.8241 & 18.32 & \textbf{80.13} \\
\cmidrule(lr){2-7}
 & 50  & 0.2467 & 0.6998 & 0.8273 & 18.32 & 87.32 \\
\cmidrule(lr){2-7}
 & 100 & \textbf{0.2478} & \textbf{0.7008} & \textbf{0.8281} & 18.32 & 102.0 \\
\bottomrule
\end{tabular}
\label{tab:chunk_ablation}
\end{table}
\begin{table}[t]
\centering
\caption{Component-wise latency (s) on the RedKitchen scene from the 7Scenes~\cite{7scenes}}
\setlength{\tabcolsep}{5pt}
\resizebox{0.8\linewidth}{!}{%
\begin{tabular}{c c c c c c c}
\toprule
Frames & Sim. Matrix & Partitioning & Alignment & Transformer & Total \\
\midrule
300 & 0.0036 & 0.0475 & 0.0017 & 27.685 & 27.753 \\
\midrule
500 & 0.0071 & 0.0938 & 0.0158 & 49.776 & 49.893 \\
\midrule
1000 & 0.0098 & 0.3426 & 0.0286 & 91.735 & 92.116 \\
\bottomrule
\end{tabular}\label{tab:time}
}
\end{table}

\paragraph{\textbf{\textup{Component-wise Inference Time Analysis.}}}
We measure the latency of VGGT~\cite{VGGT} + our framework on Redkitchen scene from the 7Scenes~\cite{7scenes} pose estimation task. As shown in Tab.~\ref{tab:time}, similarity computation, partitioning, and alignment introduce only a small overhead compared to the VGGT inference time. Even for 1000 frames, these additional steps take less than a second in total, while the runtime is dominated by the transformer inference. This indicates that our framework improves view organization with negligible additional computational cost.
\section{Conclusion}
\label{sec:conclusion}
In this paper, we analyze the impact of view redundancy in geometry transformers and show that increasing the number of views without sufficient viewpoint diversity can degrade reconstruction performance by diffusing attention across similar observations. 
Motivated by this insight,
we propose a training-free, plug-and-play framework that reorganizes input views into diverse chunks for more effective multi-view reasoning. By combining diversity-aware partitioning with soft pose propagation,
our method improves camera pose estimation, multi-view depth prediction, and 3D reconstruction accuracy while reducing memory usage and inference latency. 
The proposed framework can be seamlessly integrated with existing VGGT variants, providing a practical solution for scalable long-sequence multi-view reconstruction.

\subsection*{Acknowledgement}
This work was supported by the AI Graduate School Program at POSTECH (RS-2019-II191906 (5\%)), the NRF grants (RS-2025-24535146(10\%), RS-2026-25491789), and the IITP grants (RS-2022-II220926 (30\%), RS-2024-00457882 (20\%), RS-2026-25518317 (35\%) funded by MSIT, Korea.

\bibliographystyle{splncs04}
\bibliography{main}

@String(CVPR  = {IEEE Conf. Comput. Vis. Pattern Recog.})

@String(ICCV  = {Int. Conf. Comput. Vis.})

@String(ECCV  = {Eur. Conf. Comput. Vis.})

@String(NeurIPS = {Adv. Neural Inform. Process. Syst.})

@String(ICLR  = {Int. Conf. Learn. Represent.})

@String(TMLR  = {Trans. Mach. Learn Res.})

@String(CVPR  = {CVPR})

@String(ICCV  = {ICCV})

@String(ECCV  = {ECCV})

@String(NeurIPS = {NeurIPS})

@String(ICLR  = {ICLR})

@String(TMLR  = {TMLR})

@InProceedings{VGGT,
    author    = {Wang, Jianyuan and Chen, Minghao and Karaev, Nikita and Vedaldi, Andrea and Rupprecht, Christian and Novotny, David},
    title     = {VGGT: Visual Geometry Grounded Transformer},
    booktitle = {CVPR},
    year      = {2025},
}

@InProceedings{litevggt,
    author    = {Shu, Zhijian and Lin, Cheng and Xie, Tao and Yin, Wei and Li, Ben and Pu, Zhiyuan and Li, Weize and Yao, Yao and Cao, Xun and Guo, Xiaoyang and Long, Xiao-Xiao},
    title     = {LiteVGGT: Boosting Vanilla VGGT via Geometry-aware Cached Token Merging},
    booktitle = {CVPR},
    month     = {June},
    year      = {2026},
    pages     = {36422-36432}
}

@article{fastervggt,
  title={Faster vggt with block-sparse global attention},
  author={Wang, Chung-Shien Brian and Schmidt, Christian and Piekenbrinck, Jens and Leibe, Bastian},
  journal={arXiv preprint arXiv:2509.07120},
  year={2025}
}

@inproceedings{bonn,
  title={ReFusion: 3D reconstruction in dynamic environments for RGB-D cameras exploiting residuals},
  author={Palazzolo, Emanuele and Behley, Jens and Lottes, Philipp and Giguere, Philippe and Stachniss, Cyrill},
  booktitle={IROS},
  year={2019},
}

@INPROCEEDINGS{7scenes,
  author={Shotton, Jamie and Glocker, Ben and Zach, Christopher and Izadi, Shahram and Criminisi, Antonio and Fitzgibbon, Andrew},
  booktitle={CVPR}, 
  title={Scene Coordinate Regression Forests for Camera Relocalization in RGB-D Images}, 
  year={2013},
}

@inproceedings{nrgbd,
  title={Neural rgb-d surface reconstruction},
  author={Azinovi{\'c}, Dejan and Martin-Brualla, Ricardo and Goldman, Dan B and Nie{\ss}ner, Matthias and Thies, Justus},
  booktitle={CVPR},
  year={2022}
}

@article{dino,
  title={DINOv2: Learning Robust Visual Features without Supervision},
  author={Oquab, Maxime and Darcet, Timoth{\'e}e and Moutakanni, Th{\'e}o and Vo, Huy and Szafraniec, Marc and Khalidov, Vasil and Fernandez, Pierre and Haziza, Daniel and Massa, Francisco and El-Nouby, Alaaeldin and others},
  journal={TMLR},
  year={2024}
}

@inproceedings{dpt,
  title={Vision transformers for dense prediction},
  author={Ranftl, Ren{\'e} and Bochkovskiy, Alexey and Koltun, Vladlen},
  booktitle={ICCV},
  year={2021}
}

@inproceedings{vggtlong,
  title={VGGT-Long: Chunk it, Loop it, Align it -- Pushing VGGT's Limits on Kilometer-scale Long RGB Sequences},
  author={Deng, Kai and Ti, Zexin and Xu, Jiawei and Yang, Jian and Xie, Jin},
  booktitle={ICRA},
  year={2026}
}

@article{swiftvggt,
  title={SwiftVGGT: A Scalable Visual Geometry Grounded Transformer for Large-Scale Scenes},
  author={Lee, Jungho and Lee, Minhyeok and Yang, Sunghun and Kang, Minseok and Lee, Sangyoun},
  journal={arXiv preprint arXiv:2511.18290},
  year={2025}
}

@InProceedings{flashvggt,
    author    = {Wang, Zipeng and Xu, Dan},
    title     = {FlashVGGT: Efficient and Scalable Visual Geometry Transformers with Compressed Descriptor Attention},
    booktitle = {CVPR},
    month     = {June},
    year      = {2026},
    pages     = {21826-21835}
}

@InProceedings{httm,
    author    = {Wang, Weitian and Meiner, Lukas and Shubham, Rai and De La Parra, Cecilia and Kumar, Akash},
    title     = {HTTM: Head-wise Temporal Token Merging for Faster VGGT},
    booktitle = {CVPR},
    month     = {June},
    year      = {2026},
    pages     = {26379-26388}
}

@inproceedings{fastvggt,
title={Fast{VGGT}: Fast Visual Geometry Transformer},
author={You Shen and Zhipeng Zhang and Yansong Qu and Xiawu Zheng and Jiayi Ji and Shengchuan Zhang and Liujuan Cao},
booktitle={ICLR},
year={2026},
url={https://openreview.net/forum?id=asl8NJlIMe}
}

@inproceedings{mvp,
title={Multi-view Pyramid Transformer: Look Coarser to See Broader
},
author={Gyeongjin Kang and Seungkwon Yang and Seungtae Nam and Younggeun Lee and Jungwoo Kim and Eunbyung Park},
booktitle={CVPR},
year={2026},
}

@InProceedings{avggt,
    author    = {Sun, Xianbing and Zhu, Zhikai and Lou, Zhengyu and Yang, Bo and Tang, Jinyang and Zhang, Liqing and Wang, He and Zhang, Jianfu},
    title     = {AVGGT: Rethinking Global Attention for Accelerating VGGT},
    booktitle = {CVPR},
    month     = {June},
    year      = {2026},
    pages     = {251-260}
}

@ARTICLE{KL,
  author={Kernighan, B. W. and Lin, S.},
  journal={The Bell System Technical Journal}, 
  title={An efficient heuristic procedure for partitioning graphs}, 
  year={1970},
  volume={49},
  number={2},
  pages={291-307},
}

@article{tanks,
    author    = {Arno Knapitsch and Jaesik Park and Qian-Yi Zhou and Vladlen Koltun},
    title     = {Tanks and Temples: Benchmarking Large-Scale Scene Reconstruction},
    journal   = {ACM Transactions on Graphics},
    volume    = {36},
    number    = {4},
    year      = {2017},
}

@inproceedings{scannet,
  title={Scannet: Richly-annotated 3d reconstructions of indoor scenes},
  author={Dai, Angela and Chang, Angel X and Savva, Manolis and Halber, Maciej and Funkhouser, Thomas and Nie{\ss}ner, Matthias},
  booktitle={CVPR},
  year={2017}
}

@inproceedings{glomap,
    author={Pan, Linfei and Barath, Daniel and Pollefeys, Marc and Sch\"{o}nberger, Johannes Lutz},
    title={{Global Structure-from-Motion Revisited}},
    booktitle={ECCV},
    year={2024},
}

@inproceedings{colmap,
    author={Sch\"{o}nberger, Johannes Lutz and Frahm, Jan-Michael},
    title={Structure-from-Motion Revisited},
    booktitle={CVPR},
    year={2016},
}

@inproceedings{dust3r,
  title={Dust3r: Geometric 3d vision made easy},
  author={Wang, Shuzhe and Leroy, Vincent and Cabon, Yohann and Chidlovskii, Boris and Revaud, Jerome},
  booktitle={CVPR},
  year={2024}
}

@inproceedings{mast3r,
  title={Grounding image matching in 3d with mast3r},
  author={Leroy, Vincent and Cabon, Yohann and Revaud, J{\'e}r{\^o}me},
  booktitle={ECCV},
  year={2024},
}

@article{nbv_survey,
  author  = {Scott, William R. and Roth, Gerhard and Rivest, Jean-Fran\c{c}ois},
  title   = {View Planning for Automated Three-Dimensional Object Reconstruction and Inspection},
  journal = {ACM Computing Surveys},
  year    = {2003},
  volume  = {35},
  number  = {1},
  pages   = {64--96},
}

@INPROCEEDINGS{nbv,
  author={Connolly, C.},
  booktitle={ICRA}, 
  title={The determination of next best views}, 
  year={1985},}

@inproceedings{fast3r,
  title={Fast3r: Towards 3d reconstruction of 1000+ images in one forward pass},
  author={Yang, Jianing and Sax, Alexander and Liang, Kevin J and Henaff, Mikael and Tang, Hao and Cao, Ang and Chai, Joyce and Meier, Franziska and Feiszli, Matt},
  booktitle={CVPR},
  year={2025}
}

@article{procrustes,
    title={A Generalized Solution of the Orthogonal Procrustes Problem},
    volume={31},
    number={1},
    journal={Psychometrika},
    author={Schönemann, Peter H.},
    year={1966},
    pages={1–10}
}

@inproceedings{croco,
 author = {Weinzaepfel, Philippe and Leroy, Vincent and Lucas, Thomas and BR\'{E}GIER, Romain and Cabon, Yohann and ARORA, Vaibhav and Antsfeld, Leonid and Chidlovskii, Boris and Csurka, Gabriela and Revaud, Jerome},
 booktitle = {NeurIPS},
 title = {CroCo: Self-Supervised Pre-training for 3D Vision Tasks by Cross-View Completion},
 year = {2022}
}

@inproceedings{tome,
      title={Token Merging: Your ViT But Faster}, 
      author={Daniel Bolya and Cheng-Yang Fu and Xiaoliang Dai and Peizhao Zhang and Christoph Feichtenhofer and Judy Hoffman},
      booktitle = {ICLR},
      year={2023}
}

@InProceedings{vggsfm,
  author    = {Jianyuan Wang and Nikita Karaev and Christian Rupprecht and David Novotny},
  title     = {VGGSfM: Visual Geometry Grounded Deep Structure From Motion},
  booktitle = {CVPR},
  year      = {2024}
}

@inproceedings{pi3,
title={{$\pi$\textsuperscript{3}}: Permutation-Equivariant Visual Geometry Learning},
author={Wang, Yifan and Zhou, Jianjun and Zhu, Haoyi and Chang, Wenzheng and Zhou, Yang and Li, Zizun and Chen, Junyi and Pang, Jiangmiao and Shen, Chunhua and He, Tong},
booktitle={ICLR},
year={2026},
}

@article{splatt3r,
      title={Splatt3R: Zero-shot Gaussian Splatting from Uncalibrated Image Pairs}, 
      author={Brandon Smart and Chuanxia Zheng and Iro Laina and Victor Adrian Prisacariu},
      journal={arXiv preprint	arXiv:2408.13912},
      year={2024},
}

@inproceedings{lightglue,
  author    = {Philipp Lindenberger and
               Paul-Edouard Sarlin and
               Marc Pollefeys},
  title     = {{LightGlue: Local Feature Matching at Light Speed}},
  booktitle = {ICCV},
  year      = {2023}
}

@inproceedings{heritage,
  title={Neural {3D} Reconstruction in the Wild},
  author={Sun, Jiaming and Chen, Xi and Wang, Qianqian and Li, Zhengqi and Averbuch-Elor, Hadar and Zhou, Xiaowei and Snavely, Noah},
  booktitle={SIGGRAPH},
  year={2022}
}

@inproceedings{must3r,
  title={Must3r: Multi-view network for stereo 3d reconstruction},
  author={Cabon, Yohann and Stoffl, Lucas and Antsfeld, Leonid and Csurka, Gabriela and Chidlovskii, Boris and Revaud, Jerome and Leroy, Vincent},
  booktitle={CVPR},
  pages={1050--1060},
  year={2025}
}

@inproceedings{spann3r,
  title={3d reconstruction with spatial memory},
  author={Wang, Hengyi and Agapito, Lourdes},
  booktitle={3DV},
  pages={78--89},
  year={2025},
  organization={IEEE}
}

@inproceedings{gds,
  title={Close, but not there: Boosting geographic distance sensitivity in visual place recognition},
  author={Izquierdo, Sergio and Civera, Javier},
  booktitle={ECCV},
  pages={240--257},
  year={2024},
  organization={Springer}
}

@inproceedings{tum,
 author = {J. Sturm and N. Engelhard and F. Endres and W. Burgard and D. Cremers},
 title = {A Benchmark for the Evaluation of RGB-D SLAM Systems},
 booktitle = {IROS},
 year = {2012},
 month = {Oct.},
}

@inproceedings{megaloc,
  title={Megaloc: One retrieval to place them all},
  author={Berton, Gabriele and Masone, Carlo},
  booktitle={CVPR Workshop},
  pages={2861--2867},
  year={2025}
}

@InProceedings{salad,
    author    = {Izquierdo, Sergio and Civera, Javier},
    title     = {Optimal Transport Aggregation for Visual Place Recognition},
    booktitle = {CVPR},
    month     = {June},
    year      = {2024},
}

@inproceedings{hess,
  title={HeSS: Head sensitivity score for sparsity redistribution in VGGT},
  author={Kim, Yongsung and Song, Wooseok and Lew, Jaihyun and Hwangbo, Hun and Lee, Jaehoon and Yoon, Sungroh},
  booktitle={CVPR},
  pages={36509--36517},
  year={2026}
}

@inproceedings{superglue,
  title={Superglue: Learning feature matching with graph neural networks},
  author={Sarlin, Paul-Edouard and DeTone, Daniel and Malisiewicz, Tomasz and Rabinovich, Andrew},
  booktitle={CVPR},
  pages={4938--4947},
  year={2020}
}

@inproceedings{edm,
  title={EDM: Efficient Deep Feature Matching},
  author={Li, Xi and Rao, Tong and Pan, Cihui},
  booktitle={ICCV},
  pages={26198--26208},
  year={2025}
}

@InProceedings{loftr,
    author    = {Sun, Jiaming and Shen, Zehong and Wang, Yuang and Bao, Hujun and Zhou, Xiaowei},
    title     = {LoFTR: Detector-Free Local Feature Matching With Transformers},
    booktitle = {CVPR},
    month     = {June},
    year      = {2021},
    pages     = {8922-8931}
}

@inproceedings{gennbv,
  title={Gennbv: Generalizable next-best-view policy for active 3d reconstruction},
  author={Chen, Xiao and Li, Quanyi and Wang, Tai and Xue, Tianfan and Pang, Jiangmiao},
  booktitle={CVPR},
  pages={16436--16445},
  year={2024}
}

@inproceedings{popnbv,
  title={Pop-gs: Next best view in 3d-gaussian splatting with p-optimality},
  author={Wilson, Joey and Almeida, Marcelino and Mahajan, Sachit and Labrie, Martin and Ghaffari, Maani and Ghasemalizadeh, Omid and Sun, Min and Kuo, Cheng-Hao and Sen, Arnab},
  booktitle={CVPR},
  pages={3646--3655},
  year={2025}
}

@article{orbkf,
  title={ORB-SLAM: A versatile and accurate monocular SLAM system},
  author={Mur-Artal, Raul and Montiel, Jose Maria Martinez and Tardos, Juan D},
  journal={IEEE transactions on robotics},
  volume={31},
  number={5},
  pages={1147--1163},
  year={2015},
  publisher={IEEE}
}

@article{adaptivekf,
  title={Adaptive Keyframe Selection for Scalable 3D Scene Reconstruction in Dynamic Environments},
  author={Jha, Raman and Zhou, Yang and Loianno, Giuseppe},
  journal={arXiv preprint arXiv:2510.23928},
  year={2025}
}

@inproceedings{splatamkf,
  title={Splatam: Splat track \& map 3d gaussians for dense rgb-d slam},
  author={Keetha, Nikhil and Karhade, Jay and Jatavallabhula, Krishna Murthy and Yang, Gengshan and Scherer, Sebastian and Ramanan, Deva and Luiten, Jonathon},
  booktitle={Proceedings of the IEEE/CVF conference on computer vision and pattern recognition},
  pages={21357--21366},
  year={2024}
}
\clearpage
\begin{center}
    {\bfseries\huge Supplementary Materials} \\[6mm]
    {\large \textit{Diversity-aware View Partitioning for Scalable VGGT}}
\end{center}
\noindent\rule[0.5ex]{\textwidth}{0.4pt}

\appendix
\setcounter{figure}{5}   
\setcounter{table}{11}   

\section{Additional Implementation Details}
\label{sec:supp_imple}

\subsection{Dataset Details}

\paragraph{\textbf{\textup{7Scenes~\cite{7scenes}}}}
A standard indoor RGB-D relocalization benchmark consisting of seven small-scale scenes (Chess, Fire, Heads, Office, Pumpkin, Red Kitchen, Stairs) captured with a Kinect sensor.
Each scene provides ground-truth camera poses and depth maps, covering room-scale environments with repetitive textures.

\paragraph{\textbf{\textup{NRGBD~\cite{nrgbd}}}}
An indoor RGB-D dataset designed for neural surface reconstruction evaluation, containing synthetic and real scenes with ground-truth meshes.
Sequences are relatively dense ($\sim$1500 frames), providing challenging benchmarks for methods that must maintain geometric consistency over extended observations.

\paragraph{\textbf{\textup{ScanNet-50~\cite{scannet}}}}
A large-scale indoor RGB-D dataset containing over 1,500 reconstructed scenes from diverse real-world environments such as apartments, offices, and bathrooms.
Following the evaluation protocol of FastVGGT~\cite{fastvggt}, we uniformly sample 50 sequences from the full dataset for efficient yet representative evaluation.

\paragraph{\textbf{\textup{Bonn~\cite{bonn}}}}
An indoor RGB-D dataset specifically designed for dynamic scene evaluation, featuring scenes with moving people and objects.
It provides ground-truth depth and camera trajectories, and is commonly used for benchmarking video depth estimation under dynamic conditions.


\paragraph{\textbf{\textup{Heritage-Recon~\cite{heritage}}}}
A large-scale outdoor benchmark consisting of four cultural heritage landmarks with ground-truth LiDAR scans from Open Heritage 3D, paired with Internet-sourced images and SfM~\cite{colmap} camera poses.
The images are inherently unordered with significant variation in illumination and viewpoint.
Heritage-Recon is evaluated in Sec.~\ref{sec:supp_unordered}.

\paragraph{\textbf{\textup{TUM-RGBD~\cite{tum}}}}
An indoor RGB-D benchmark widely used for evaluating SLAM and camera trajectory estimation, captured with a handheld Kinect sensor across diverse office and household environments.
It provides accurate ground-truth camera trajectories from an external motion-capture system, and its long, temporally ordered sequences make it a standard benchmark for assessing long-sequence pose estimation and drift.

\paragraph{\textbf{\textup{Tanks\&Temples~\cite{tanks}}}}
A large-scale outdoor and indoor benchmark for image-based 3D reconstruction, comprising high-resolution video sequences of real-world scenes captured under realistic conditions.
It provides ground-truth geometry acquired with a high-precision laser scanner, and is commonly used for benchmarking reconstruction quality at the object and scene scale via F-score evaluation.

\subsection{Model Parameters}
\label{subsec:supp_imple_mp}
We report the model configurations used in our experiments, including optimization iterations, and other settings for each models.

\paragraph{\textbf{\textup{More Parameter Ablation}}}
We conduct ablation studies on key hyperparameters by sweeping each over a range of values while fixing all others.
Specifically, we ablate $\gamma$, the softmax temperature in soft pose propagation; $\epsilon$, which controls the linear combination weight of the soft pose term in the KL~\cite{KL}-based local search score function; $\tau$, the scale-normalization factor in the spatial dispersion matrix (Eq.~(\ref{eq:pose_weight})); and the number of local search iterations (see Sec.~\ref{subsec:pseudo_pose}).
This analysis validates our default choices and examines their impact on performance.
All experiments are evaluated on the 7Scenes~\cite{7scenes} dataset with an input frame length of $N{=}500$ and partition size $C{=}50$. The results are shown in Tab.~\ref{tab:hp_ablation}. Since the hyperparameters exhibit minimal interdependence, we ablate each one separately rather than performing a joint grid search.
\begin{table}[H]
\centering
\caption{Hyperparameter ablation on 7Scenes~\cite{7scenes}.}
\resizebox{0.9\linewidth}{!}{%
\setlength{\tabcolsep}{6pt}
\renewcommand{\arraystretch}{1.3}
\begin{tabular}{cc|cc|cc|cc}
\specialrule{0.8pt}{0pt}{0pt}
\multicolumn{1}{c}{$\gamma$} & \multicolumn{1}{c|}{AUC@3$\uparrow$} &
\multicolumn{1}{c}{$\epsilon$} & \multicolumn{1}{c|}{AUC@3$\uparrow$} &
\multicolumn{1}{c}{Iterations} & \multicolumn{1}{c|}{AUC@3$\uparrow$} &
\multicolumn{1}{c}{$\tau_{\text{mult}}$} & \multicolumn{1}{c}{AUC@3$\uparrow$} \\
\hline
1     & 0.2489 & 1     & 0.2496 & 20 & 0.2483 & 0.25 & 0.2500 \\
0.1   & 0.2485 & 0.1   & 0.2492 & 10 & 0.2482 & 0.5  & 0.2489 \\
0.01  & 0.2477 & 0.01  & 0.2496 & 5  & \textbf{0.2486} & 1    & \textbf{0.2502} \\
0.005 & 0.2481 & 0.005 & \textbf{0.2502} & 0  & 0.2248 & 2    & 0.2483 \\
0.001 & \textbf{0.2502} & 0.001 & 0.2486 & -- & --     & 4    & 0.2490 \\
\specialrule{0.8pt}{0pt}{0pt}
\end{tabular}%
}
\label{tab:hp_ablation}
\end{table}
As shown in Tab.~\ref{tab:hp_ablation}, we observe that smaller values of $\gamma$ yield better performance, with $\gamma{=}0.001$ achieving the highest AUC@3, suggesting that a sharper softmax distribution in pose propagation leads to more precise partitioning.
For $\epsilon$, performance peaks at $\epsilon{=}0.005$, indicating that a moderate soft pose contribution in the score function provides the best balance between geometric and visual diversity cues.
For $\tau$, we sweep multiplicative factors of the per-sequence median pairwise distance $\tau_{\text{med}}$, with the best AUC@3 at a factor of $1$; performance varies only narrowly across all settings, indicating low sensitivity to $\tau$ and supporting our use of the median distance for scale normalization.
For the number of local search iterations, performance improves up to 5 iterations and \emph{saturates beyond this point}, with diminishing returns at 10 and 20 iterations.
Based on these results, we adopt $\mathbf{\gamma{=}0.001}$, $\mathbf{\epsilon{=}0.005}$, $\boldsymbol{\tau{=}\tau_{\text{med}}}$, and \textbf{5 local search iterations} as our default configuration

\subsection{Visual Feature Source for Dissimilarity}
\label{sec:delta_source}

Our framework measures visual dissimilarity $\delta$ from per-frame DINOv2~\cite{dino} embeddings (Eq.~(\ref{eq:vis_dis})), obtained for free from the VGGT~\cite{VGGT} backbone with no extra forward passes.
A natural question is whether dedicated visual place recognition (VPR) descriptors, explicitly trained for viewpoint and appearance variation, yield stronger partitions.
To examine this, we replace the DINOv2-based $\delta$ with two representative VPR descriptors, SALAD~\cite{salad} and MegaLoc~\cite{megaloc}, as drop-in alternatives, and evaluate camera pose estimation on NRGBD~\cite{nrgbd}.

\begin{table}[H]
\centering
\caption{$\delta$-source on NRGBD~\cite{nrgbd}.}
\resizebox{0.6\linewidth}{!}{%
\setlength{\tabcolsep}{8pt}
\renewcommand{\arraystretch}{1.3}
\begin{tabular}{lcc}
\specialrule{0.8pt}{0pt}{0pt}
$\delta$ source & AUC@3$\uparrow$ & L(s)$\Delta$ \\
\specialrule{0.4pt}{2pt}{2pt}
DINOv2~\cite{dino} (ours) & \textbf{0.7844} $\pm$ 0.001 & -- \\
SALAD~\cite{salad} & 0.7799 $\pm$ 0.001 & \textcolor{red}{$+8.03$} \\
MegaLoc~\cite{megaloc} & 0.7821 $\pm$ 0.002 & \textcolor{red}{$+6.60$} \\
\specialrule{0.8pt}{0pt}{0pt}
\end{tabular}%
}
\label{tab:delta-source}
\end{table}

As shown in Tab.~\ref{tab:delta-source}, all three feature sources yield comparable AUC@3, with differences below $0.01$.
This indicates that the partitioning objective is largely insensitive to the specific choice of descriptor, as long as it captures coarse appearance-level dissimilarity between frames.
Crucially, however, both SALAD and MegaLoc require a separate forward pass over all input frames to extract their descriptors, adding $+8.03$s and $+6.60$s of latency respectively, whereas our DINOv2 features are already computed within the VGGT backbone and incur no extra cost.
Since the dedicated descriptors provide no measurable accuracy gain while increasing latency, we adopt the reused DINOv2 embeddings as our default $\delta$ source, preserving the training-free and overhead-free nature of our framework.

\subsection{Number of Anchor Frames}
\label{sec:anchor_count}
Our framework aligns per-chunk predictions through SE(3) registration using a single shared anchor frame (Sec.~\ref{subsec:all_inference}).
A natural question is whether using more anchor frames $A$ improves the global consistency of the merged reconstruction.
We ablate $A$ on the three longest TUM-RGBD~\cite{tum} sequences in Tab.~\ref{tab:anchor}.
\begin{table}[H]
\centering
\caption{Anchor-count ablation on the three longest TUM-RGBD~\cite{tum} sequences.}
\resizebox{0.55\linewidth}{!}{%
\setlength{\tabcolsep}{6pt}
\renewcommand{\arraystretch}{1.3}
\begin{tabular}{cccc}
\specialrule{0.8pt}{0pt}{0pt}
$A$ & ATE $\downarrow$ (cm) & CoV($s_k$) $\downarrow$ & $L(s)\downarrow$ \\
\hline
1  & 2.07          & 0.0052          & \textbf{478} \\
5  & \textbf{1.97} & 0.0012          & 511 \\
20 & \textbf{1.97} & \textbf{0.0008} & 615 \\
\specialrule{0.8pt}{0pt}{0pt}
\end{tabular}%
}
\label{tab:anchor}
\end{table}
Increasing $A$ from $1$ to $5$ and $20$ yields only marginal changes in ATE while steadily increasing latency, indicating that a single anchor is sufficient for accurate alignment.
To verify that this does not introduce scale inconsistency, we additionally report the coefficient of variation (CoV${=}\sigma/\mu$) of the per-chunk Sim(3) scales, which captures scale discrepancies that SE(3) alignment cannot correct.
The single-anchor CoV is already small and decreases only slightly with more anchors, confirming that our chunks are scale-consistent and exhibit no meaningful chunk-level scale bias.
Based on this analysis, we adopt $A{=}1$ as our default.
\section{Stability of Partitioning}
\label{sec:supp_eval_stab}

Our method initializes partitions randomly and runs a fixed number of optimization iterations (see Sec.~\ref{subsec:supp_imple_mp}), which introduces inherent stochasticity into the partitioning process.
To assess whether this randomness affects reconstruction quality, we conduct two complementary stability analyses using camera pose estimation on the 7Scenes~\cite{7scenes} and NRGBD~\cite{nrgbd} datasets.
For each configuration, we repeat the experiment across five random seeds $\{42, 123, 456, 789, 1024\}$ and report the mean and standard deviation of all metrics.

\subsection{Randomness Comparison}
We first evaluate the sensitivity to random initialization by fixing the input frame length to $N{=}500$ and running our method across five random seeds for each partition size $C \in \{25, 50, 100\}$.
This measures how much the reconstruction outputs vary solely due to different random initializations under identical external conditions.
As shown in \cref{tab:stab_1}, the standard deviation of AUC across seeds remains below $0.0012$ for all thresholds and configurations on both datasets, indicating that our randomized partitioning algorithm introduces negligible variance in the final pose accuracy.

\begin{table}[H]
\caption{Partitioning Randomness Comparison.}
\centering
\scriptsize
\setlength{\tabcolsep}{3pt}
\begin{tabular}{c c ccc}
\toprule
Dataset & Partition Size & AUC@3$\uparrow$ & AUC@15$\uparrow$ & AUC@30$\uparrow$ \\
\midrule
\multirow{3}{*}[-1.5ex]{7Scenes~\cite{7scenes}} & 25 & 0.2402{\tiny$\pm$0.0012} & 0.6957{\tiny$\pm$0.0006} & 0.8246{\tiny$\pm$0.0003} \\
\cmidrule(lr){2-5}
& 50 & 0.2481{\tiny$\pm$0.0011} & 0.7009{\tiny$\pm$0.0005} & 0.8282{\tiny$\pm$0.0004} \\
\cmidrule(lr){2-5}
& 100 & 0.2488{\tiny$\pm$0.0003} & 0.7015{\tiny$\pm$0.0003} & 0.8288{\tiny$\pm$0.0002} \\
\midrule
\multirow{3}{*}[-1.5ex]{NRGBD~\cite{nrgbd}} & 25 & 
0.7734{\tiny$\pm$0.0007} & 0.9359{\tiny$\pm$0.0004} & 0.9659{\tiny$\pm$0.0002} \\
\cmidrule(lr){2-5}
& 50 & 0.7838{\tiny$\pm$0.0008} & 0.9400{\tiny$\pm$0.0003} & 0.9685{\tiny$\pm$0.0002} \\
\cmidrule(lr){2-5}
& 100 & 0.7806{\tiny$\pm$0.0005} & 0.9394{\tiny$\pm$0.0002} & 0.9683{\tiny$\pm$0.0001} \\
\bottomrule
\end{tabular}
\label{tab:stab_1}
\end{table}

\subsection{Per-Partition Uniformity}
We then examine whether our partitioning algorithm produces partitions of uniform quality.
For each sequence (7Scenes~\cite{7scenes}) or scene (NRGBD~\cite{nrgbd}), we compute the standard deviation of per-partition AUC across all partitions within a single run, then average this value over all sequences/scenes.
We report $\sigma${\tiny$\pm$std} where the mean is taken over five seeds, confirming that partition uniformity is consistent regardless of random initialization.
As shown in \cref{tab:stab_2}, all configurations exhibit $\sigma < 0.02$ across every threshold, indicating that partitions achieve similar reconstruction quality without concentrating easy or hard frames.
Moreover, larger partition sizes yield lower $\sigma$, as each partition covers more frames and thus better represents the overall distribution.

\begin{table}[H]
\caption{Per-partition Uniformity Comparison}
\centering
\scriptsize
\setlength{\tabcolsep}{3pt}
\begin{tabular}{c c c ccc}
\toprule
Dataset & Frames & Partition Size & $\sigma$(AUC@3)$\downarrow$ & $\sigma$(AUC@15)$\downarrow$ & $\sigma$(AUC@30)$\downarrow$ \\
\midrule
\multirow{6}{*}{7Scenes~\cite{7scenes}} 
& \multirow{3}{*}{300} & 25 & 0.0190{\tiny$\pm$0.0012} & 0.0148{\tiny$\pm$0.0005} & 0.0100{\tiny$\pm$0.0007} \\
& & 50 & 0.0105{\tiny$\pm$0.0006} & 0.0082{\tiny$\pm$0.0005} & 0.0056{\tiny$\pm$0.0004} \\
& & 100 & 0.0055{\tiny$\pm$0.0011} & 0.0046{\tiny$\pm$0.0006} & 0.0031{\tiny$\pm$0.0004} \\
\cmidrule(lr){2-6}
& \multirow{3}{*}{500} & 25 & 0.0185{\tiny$\pm$0.0006} & 0.0145{\tiny$\pm$0.0003} & 0.0098{\tiny$\pm$0.0003} \\
& & 50 & 0.0111{\tiny$\pm$0.0006} & 0.0086{\tiny$\pm$0.0006} & 0.0058{\tiny$\pm$0.0005} \\
& & 100 & 0.0066{\tiny$\pm$0.0005} & 0.0051{\tiny$\pm$0.0003} & 0.0034{\tiny$\pm$0.0001} \\
\midrule
\multirow{6}{*}{NRGBD~\cite{nrgbd}} 
& \multirow{3}{*}{300} & 25 & 0.0210{\tiny$\pm$0.0009} & 0.0074{\tiny$\pm$0.0006} & 0.0047{\tiny$\pm$0.0002} \\
& & 50 & 0.0148{\tiny$\pm$0.0000} & 0.0050{\tiny$\pm$0.0000} & 0.0029{\tiny$\pm$0.0000} \\
& & 100 & 0.0082{\tiny$\pm$0.0014} & 0.0028{\tiny$\pm$0.0003} & 0.0015{\tiny$\pm$0.0002} \\
\cmidrule(lr){2-6}
& \multirow{3}{*}{500} & 25 & 0.0200{\tiny$\pm$0.0002} & 0.0067{\tiny$\pm$0.0003} & 0.0038{\tiny$\pm$0.0004} \\
& & 50 & 0.0126{\tiny$\pm$0.0010} & 0.0043{\tiny$\pm$0.0002} & 0.0023{\tiny$\pm$0.0000} \\
& & 100 & 0.0084{\tiny$\pm$0.0007} & 0.0027{\tiny$\pm$0.0003} & 0.0014{\tiny$\pm$0.0001} \\
\bottomrule
\end{tabular}
\label{tab:stab_2}
\end{table}

\section{Soft Pose Propagation Analysis}
\label{sec:softpose_analysis}

\setlength{\intextsep}{7pt}
\begin{wrapfigure}[12]{r}{0.45\columnwidth}
\centering
\includegraphics[width=\linewidth]{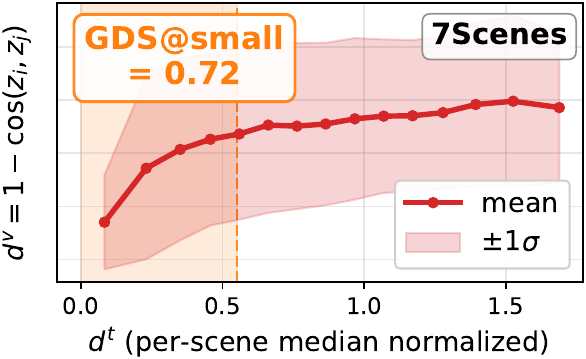}
\caption{Visual--spatial correlation on 7Scenes~\cite{7scenes}.}
\label{fig:gds_curve}
\end{wrapfigure}

Soft pose propagation (Sec.~\ref{subsec:pseudo_pose}) is designed as a \emph{coarse spatial cue} for partitioning rather than a metric pose estimator.
To assess the quality of this cue directly, we measure the median relative camera-center direction error for local frame pairs on NRGBD~\cite{nrgbd}, which is only $4.21^\circ$, confirming that the propagated poses capture useful local spatial trends.
To further verify that visual similarity is a reliable proxy for spatial proximity, we compare the DINOv2~\cite{dino} descriptor distance $d^{v}_{ij}{=}1-\cos(\mathbf{z}_i,\mathbf{z}_j)$ against the ground-truth camera-center distance $d^{t}_{ij}{=}\|\mathbf{t}_i-\mathbf{t}_j\|_2$ following~\cite{gds}.
As shown in Fig.~\ref{fig:gds_curve}, the global descriptor similarity (GDS) remains above $0.5$ in the small-distance regime, indicating a reliable local visual--spatial correlation precisely in the range most relevant to chunking.
When seed affinities are nearly uniform, the propagated pose degenerates into a weak centroid cue, so partitioning is then driven mainly by visual diversity.

\section{Ablation on Visual Dissimilarity}
\begin{table}[t]
\centering
\caption{Score functions ablation on NRGBD~\cite{nrgbd}.}
\resizebox{0.5\linewidth}{!}{%
\begin{tabular}{l c l}
\toprule
Methods & CD$\downarrow$ & ($-\Delta$) \\
\midrule
Random Init.                          & 0.036 & \\
w/ Raw similarity ($s$)               & 0.243 & \textcolor[rgb]{0.55,0,0}{+ 0.207} \\
w/ Dissimilarity ($\boldsymbol{\delta}$) & 0.029 & \textcolor[rgb]{0,0.55,0}{- 0.007} \\
\bottomrule
\end{tabular}%
}
\label{tab:score_ablation}
\end{table}
We further examine the score function of the diversity-aware view partitioning. As shown in Tab.~\ref{tab:score_ablation}, visual dissimilarity ($\delta$) consistently outperforms raw cosine similarity ($s$) and the random baseline in Chamfer distance on NRGBD~\cite{nrgbd}, confirming that maximizing within-chunk visual dissimilarity is key to effective partitioning.
\section{Chunk Sparsity Analysis}
\label{sec:sparsity}
While our diversity-aware partitioning improves reconstruction by promoting viewpoint diversity within each chunk, overly sparse chunks remain inherently challenging for chunked feed-forward inference.
Maximizing diversity without bound can push the frames in a chunk too far apart, leaving the views spatially scattered and weakening the geometric cues that VGGT~\cite{VGGT} relies on for reliable multi-view reasoning.
To analyze this diversity--sparsity trade-off, we measure chunk sparsity as the mean pairwise ground-truth camera-center distance within each chunk, and report it alongside the resulting ATE under different chunk capacities $c$ on TUM-RGBD~\cite{tum} and Tanks\&Temples~\cite{tanks}.
\begin{table}[H]
\centering
\caption{Chunk sparsity analysis on TUM-RGBD~\cite{tum} and Tanks\&Temples~\cite{tanks}.}
\resizebox{0.7\linewidth}{!}{%
\setlength{\tabcolsep}{6pt}
\renewcommand{\arraystretch}{1.3}
\begin{tabular}{cccccc}
\specialrule{0.8pt}{0pt}{0pt}
Dataset & Metric & 5 & 25 & 100 & VL100$^{\dagger}$ \\
\hline
\multirow{2}{*}{TUM}  & Sparsity (m)         & 0.526 & 0.218 & 0.106          & 0.026 \\
                      & ATE (m) $\downarrow$ & 0.193 & 0.063 & \textbf{0.039} & 0.134 \\
\hline
\multirow{2}{*}{T\&T} & Sparsity (m)         & 2.309 & 0.965 & 0.471          & 0.376 \\
                      & ATE (m) $\downarrow$ & 0.601 & 0.303 & \textbf{0.267} & 0.581 \\
\specialrule{0.8pt}{0pt}{0pt}
\multicolumn{6}{@{}l@{}}{\footnotesize $^{\dagger}$VGGT-Long result with chunk size $100$.}\\
\end{tabular}%
}
\label{tab:sparsity}
\end{table}
As shown in Tab.~\ref{tab:sparsity}, smaller capacities ($c{=}5$) yield highly sparse chunks and the worst ATE, since their frames are spread too far apart to support a stable reconstruction.
At the other extreme, temporal chunking such as VGGT-Long (VL) produces very dense chunks (lowest sparsity) but packs them with redundant nearby views that offer limited viewpoint diversity, leading to noticeably higher ATE than our partitioning at a comparable chunk size.
Our default capacity lies between these two regimes: it keeps the frames within each chunk close enough to remain spatially coherent while retaining the viewpoint diversity needed for strong geometric cues, achieving the best ATE on both datasets.
This confirms that our method balances diversity against excessive sparsity rather than maximizing diversity alone, and that the chunk capacity $c$ exposes this trade-off as a single, interpretable control.

\section{Unordered Sequences Evaluations}
\label{sec:supp_unordered}
A key advantage of VGGT~\cite{VGGT} is its order-agnostic nature: 
bidirectional attention makes predictions invariant to the input ordering, 
enabling robust reconstruction from unstructured image collections.
However, extending VGGT to longer sequences often comes at the cost of this property.
VGGT-Long~\cite{vggtlong} processes the input as overlapping sequential partitions 
with Sim(3) alignment~\cite{procrustes} and loop closure, inherently assuming temporally coherent ordering.
When input frames are unordered, partition overlap becomes meaningless 
and the alignment pipeline breaks down.
Since our method is built upon VGGT's full-attention backbone, 
it naturally inherits this order-agnostic property while achieving scalability.
To verify this, we evaluate on the Heritage-Recon~\cite{heritage} dataset, 
whose images are inherently unordered Internet photo collections,
comparing our method against VGGT~\cite{VGGT} and VGGT-Long~\cite{vggtlong}.


\subsection{Results}
For fair comparison, both methods use equal partition sizes. Unlike VGGT-Long~\cite{vggtlong}, which relies on temporal coherence and requires 30 overlapping frames for Sim(3) alignment, our method needs only a single shared frame for simple SE(3) alignment. Consequently, VGGT-Long degrades on this unordered dataset (Tab.~\ref{tab:supp_unorder}), whereas our method reduces ATE by over $25\%$ and significantly boosts AUC@3. This improvement stems from reduced scale ambiguity enabled by our diversity-aware partitioning. Furthermore, consuming $38\%$ less VRAM, our approach efficiently preserves VGGT's~\cite{VGGT} order-agnostic capabilities.
\begin{table}[H]
\caption{Unordered-sequences evaluations on Heritage-Recon~\cite{heritage}}
\centering
\scriptsize
\setlength{\tabcolsep}{3pt}
\begin{tabular}{@{\hskip 4pt}lcccccc}
\toprule
\multicolumn{1}{@{\hskip 4pt}c}{Model} & ATE$\downarrow$ & RPE$_\text{trans}\downarrow$ & RPE$_\text{rot}\downarrow$ & AUC@3$\uparrow$ & AUC@30$\uparrow$ & VRAM(GB)$\downarrow$ \\
\midrule
VGGT$^*$~\cite{VGGT} & OOM & OOM & OOM & OOM & OOM & OOM\\
VGGT-Long~\cite{vggtlong} & 0.6598 & \textbf{4.020} & 2.997 & 0.3673 & 0.8082 & 37.00 \\
VGGT~\cite{VGGT} + Ours & \textbf{0.4881} & 4.031 & \textbf{2.622} & \textbf{0.5822} & \textbf{0.8960} & \textbf{22.80} \\
\bottomrule
\end{tabular}
\label{tab:supp_unorder}
\end{table}

\section{More Quantitative Results}
\label{sec:supp_quant}
We provide additional quantitative results to further demonstrate the robustness and scalability of our method.
\subsection{Camera Pose Estimations}
Tab.~\ref{tab:cam_nrgbd} details additional camera pose estimation results on NRGBD~\cite{nrgbd}. Our method consistently improves AUC metrics across all VGGT variants while significantly reducing VRAM and latency. Notably, it prevents the 1000-frame OOM failures and performance degradation afflicting baseline models.
\begin{table}[H]
\caption{Camera pose estimation on NRGBD~\cite{nrgbd}}
\centering
\scriptsize
\resizebox{\linewidth}{!}{%
\begin{tabular}{c l r ccccc}
\toprule
{Frames} & \multicolumn{2}{c}{{Model}} &
{AUC@3}$\uparrow$ & {AUC@15}$\uparrow$ & {AUC@30}$\uparrow$ &
{VRAM(GB)}$\downarrow$ & {Latency(s)}$\downarrow$ \\
\midrule

\multirow{6}{*}[-1.5ex]{300} & \quad VGGT$^*$\cite{VGGT}        &         & 0.6972 & 0.9079 & 0.9511 & 18.76 & 55.21 \\
                    & \quad VGGT$^*$\cite{VGGT}            & + Ours  & \textbf{0.7887} & \textbf{0.9423} & \textbf{0.9700} & \textbf{12.66} & \textbf{27.71} \\
\cmidrule(lr){2-8}
                    & \quad FasterVGGT\cite{fastervggt} &        & 0.3939 & 0.8306 & 0.9120 & 25.18 & 39.39 \\
                    & \quad FasterVGGT\cite{fastervggt} & + Ours & \textbf{0.4127} & \textbf{0.8466} & \textbf{0.9203} & \textbf{12.98} & \textbf{26.12} \\
\cmidrule(lr){2-8}
                    & \quad LiteVGGT\cite{litevggt}    &         & 0.5959 & 0.8444 & 0.9118 & 16.43 & 14.88 \\
                    & \quad LiteVGGT\cite{litevggt}    & + Ours  & \textbf{0.6247} & \textbf{0.8643} & \textbf{0.9244} & \textbf{10.04} & \textbf{9.95} \\
\midrule

\multirow{6}{*}[-1.5ex]{500} & \quad VGGT$^*$\cite{VGGT}        &         & 0.6625 & 0.8777 & 0.9325 & 28.41 & 150.84 \\
                    & \quad VGGT$^*$\cite{VGGT}            & + Ours  & \textbf{0.7859} & \textbf{0.9412} & \textbf{0.9692} & \textbf{16.99} & \textbf{51.32} \\
\cmidrule(lr){2-8}
                    & \quad FasterVGGT\cite{fastervggt} &        & 0.3737 & 0.7904 & 0.8801 & 40.90 & 94.00 \\
                    & \quad FasterVGGT\cite{fastervggt} & + Ours & \textbf{0.4129} & \textbf{0.8460} & \textbf{0.9196} & \textbf{15.65} & \textbf{42.83} \\
\cmidrule(lr){2-8}
                    & \quad LiteVGGT\cite{litevggt}    &         & 0.5410 & 0.7941 & 0.8510 & 25.94 & 33.34 \\
                    & \quad LiteVGGT\cite{litevggt}    & + Ours  & \textbf{0.6198} & \textbf{0.8709} & \textbf{0.9292} & \textbf{11.91} & \textbf{15.84} \\
\midrule

\multirow{6}{*}[-1.5ex]{1000} & \quad VGGT$^*$\cite{VGGT}        &         & 0.6283 & 0.8226 & 0.8817 & 69.56 & 584.20 \\
                     & \quad VGGT$^*$\cite{VGGT}           & + Ours  & \textbf{0.7859} & \textbf{0.9412} & \textbf{0.9683} & \textbf{18.31} & \textbf{93.11} \\
\cmidrule(lr){2-8}
                     & \quad FasterVGGT\cite{fastervggt} &       & $OOM$ & $OOM$ & $OOM$ & $OOM$ & $OOM$ \\
                     & \quad FasterVGGT\cite{fastervggt} & + Ours & \textbf{0.4091} & \textbf{0.8436} & \textbf{0.9178} & \textbf{18.31} & \textbf{83.83} \\
\cmidrule(lr){2-8}
                     & \quad LiteVGGT\cite{litevggt}   &         & 0.4474 & 0.7590 & 0.8252 & 48.62 & 93.91 \\
                     & \quad LiteVGGT\cite{litevggt}   & + Ours  & \textbf{0.6205} & \textbf{0.8668} & \textbf{0.9254} & \textbf{13.98} & \textbf{32.10} \\
\bottomrule
\end{tabular}%
}
\label{tab:cam_nrgbd}
\end{table}
\subsection{Multi-view Depth Estimations}
Tab.~\ref{tab:depth_full} presents additional quantitative results for multi-view depth estimation on the Bonn~\cite{bonn} dataset at 300 frames. As demonstrated, integrating our framework substantially reduces VRAM consumption and inference latency across all baseline architectures. Although we observe a marginal decrease in the $\delta_1$ accuracy metric when applying our method, a discussion and analysis regarding this slight performance gap are provided in Sec. \ref{sec:supp_fail}.
\begin{table}[H]
\caption{Multi-view Depth Estimation on Bonn~\cite{bonn}.}
\centering
\scriptsize
\resizebox{0.8\linewidth}{!}{%
\begin{tabular}{c l r cccc}
\toprule
{Frames} & \multicolumn{2}{c}{{Model}} &
{Abs Rel}$\downarrow$ & {$\bm{\delta_1}$}$\uparrow$ &
{VRAM(GB)}$\downarrow$ & {Latency(s)}$\downarrow$ \\
\midrule
\multirow{6}{*}[-1.5ex]{300} & \quad VGGT$^*$\cite{VGGT}            &        & \textbf{0.048} & \textbf{0.958} & 18.89 & 62.51 \\
                    & \quad VGGT$^*$\cite{VGGT}                & + Ours & \textbf{0.048} & 0.954          & \textbf{12.88} & \textbf{33.96} \\
\cmidrule(lr){2-7}
                    & \quad FasterVGGT\cite{fastervggt}     &        & \textbf{0.047} & \textbf{0.958} & 25.30 & 44.18 \\
                    & \quad FasterVGGT\cite{fastervggt}     & + Ours & \textbf{0.047} & 0.956          & \textbf{13.11} & \textbf{29.74} \\
\cmidrule(lr){2-7}
                    & \quad LiteVGGT\cite{litevggt}         &        & \textbf{0.056}          & \textbf{0.962} & 16.50 & 17.01 \\
                    & \quad LiteVGGT\cite{litevggt}         & + Ours & \textbf{0.056} & 0.955          & \textbf{10.04} & \textbf{11.01} \\
\midrule
\multirow{6}{*}[-1.5ex]{500} & \quad VGGT$^*$\cite{VGGT}            &        & \textbf{0.048} & \textbf{0.961} & 28.53 & 156.47 \\
                    & \quad VGGT$^*$\cite{VGGT}                & + Ours & \textbf{0.048} & 0.954          & \textbf{15.78} & \textbf{65.84} \\
\cmidrule(lr){2-7}
                    & \quad FasterVGGT\cite{fastervggt}     &        & \textbf{0.046} & \textbf{0.959} & 41.02 & 102.6 \\
                    & \quad FasterVGGT\cite{fastervggt}     & + Ours & 0.047          & 0.956          & \textbf{15.78} & \textbf{49.92} \\
\cmidrule(lr){2-7}
                    & \quad LiteVGGT\cite{litevggt}         &        & 0.057          & \textbf{0.964} & 26.01 & 35.55 \\
                    & \quad LiteVGGT\cite{litevggt}         & + Ours & \textbf{0.055} & 0.956          & \textbf{11.98} & \textbf{14.69} \\
\bottomrule
\end{tabular}%
}
\label{tab:depth_full}
\end{table}
\subsection{Multi-view 3D Reconstructions}
Tab.~\ref{tab:3d_full_1} and \ref{tab:3d_full_2} present extended 3D reconstruction results on three datasets~\cite{scannet, 7scenes, nrgbd}. Our method significantly reduces VRAM and latency, improves Chamfer Distance (CD), and resolves 1000-frame OOM failures. The marginal NC drop is analyzed in Sec.~\ref{sec:supp_fail}.
\begin{table}[H]
\caption{Multi-view 3D reconstruction on 7Scenes~\cite{7scenes} and NRGBD~\cite{nrgbd}}
\centering
\scriptsize
\setlength{\tabcolsep}{5pt}
\resizebox{\linewidth}{!}{%
\begin{tabular}{c l r ccc ccc}
\toprule
\multirow{2}{*}{Frames} & \multicolumn{2}{c}{\multirow{2}{*}{Model}} &
\multicolumn{3}{c}{7Scenes} & \multicolumn{3}{c}{NRGBD} \\
\cmidrule(lr){4-6}\cmidrule(lr){7-9}
 & \multicolumn{2}{c}{} &
CD$\downarrow$ & NC$\uparrow$ & L(s)$\downarrow$ &
CD$\downarrow$ & NC$\uparrow$ & L(s)$\downarrow$ \\
\midrule

\multirow{6}{*}[-1.5ex]{300} & \quad VGGT$^*$\cite{VGGT}              &        & 0.046 & \textbf{0.653} & 61.88 & 0.033 & 0.822 & 60.14 \\
                    & \quad VGGT$^*$\cite{VGGT}                  & + Ours & \textbf{0.043} & 0.649 & \textbf{34.13} & \textbf{0.029} & \textbf{0.859} & \textbf{32.20} \\
\cmidrule(lr){2-9}
                    & \quad FasterVGGT\cite{fastervggt}       &        & 0.046 & \textbf{0.629} & 41.88 & 0.048 & \textbf{0.759} & 42.63 \\
                    & \quad FasterVGGT\cite{fastervggt}       & + Ours & \textbf{0.043} & 0.618 & \textbf{29.24} & \textbf{0.043} & 0.755 & \textbf{29.51} \\
\cmidrule(lr){2-9}
                    & \quad LiteVGGT\cite{litevggt}           &        & 0.047 & \textbf{0.637} & 17.00 & 0.049 & \textbf{0.758} & 17.81 \\
                    & \quad LiteVGGT\cite{litevggt}           & + Ours & \textbf{0.045} & 0.624 & \textbf{10.25} & \textbf{0.042} & \textbf{0.758} & \textbf{11.07} \\

\midrule

\multirow{6}{*}[-1.5ex]{500} & \quad VGGT$^*$\cite{VGGT}              &        & 0.045 & 0.648 & 155.38 & 0.037 & 0.810 & 155.95 \\
                    & \quad VGGT$^*$\cite{VGGT}                  & + Ours & \textbf{0.042} & \textbf{0.649} & \textbf{59.04} & \textbf{0.028} & \textbf{0.851} & \textbf{58.32} \\
\cmidrule(lr){2-9}
                    & \quad FasterVGGT\cite{fastervggt}       &        & 0.046 & \textbf{0.624} & 99.57 & 0.048 & 0.742 & 103.5 \\
                    & \quad FasterVGGT\cite{fastervggt}       & + Ours & \textbf{0.043} & 0.613 & \textbf{51.69} & \textbf{0.041} & \textbf{0.743} & \textbf{49.81} \\
\cmidrule(lr){2-9}
                    & \quad LiteVGGT\cite{litevggt}           &        & 0.049 & \textbf{0.626} & 34.48 & 0.069 & 0.743 & 40.18 \\
                    & \quad LiteVGGT\cite{litevggt}           & + Ours & \textbf{0.045} & 0.624 & \textbf{17.16} & \textbf{0.049} & \textbf{0.745} & \textbf{16.84} \\

\midrule

\multirow{6}{*}[-1.5ex]{1000} & \quad VGGT$^*$\cite{VGGT}              &        & 0.061 & 0.633 & 596.11 & 0.043 & 0.806 & 596.07 \\
                    & \quad VGGT$^*$\cite{VGGT}                  & + Ours & \textbf{0.042} & \textbf{0.646} & \textbf{98.19} & \textbf{0.030} & \textbf{0.851} & \textbf{101.39} \\
\cmidrule(lr){2-9}
                    & \quad FasterVGGT\cite{fastervggt}       &        & OOM & OOM & OOM & OOM & OOM & OOM \\
                    & \quad FasterVGGT\cite{fastervggt}       & + Ours & \textbf{0.043} & \textbf{0.610} & \textbf{97.13} & \textbf{0.041} & \textbf{0.741} & \textbf{98.30} \\
\cmidrule(lr){2-9}
                    & \quad LiteVGGT\cite{litevggt}           &        & 0.063 & 0.612 & 97.11 & 0.073 & 0.715 & 99.88 \\
                    & \quad LiteVGGT\cite{litevggt}           & + Ours & \textbf{0.044} & \textbf{0.617} & \textbf{33.27} & \textbf{0.039} & \textbf{0.743} & \textbf{34.01} \\
\bottomrule
\end{tabular}%
}
\label{tab:3d_full_1}
\end{table}
\begin{table}[H]
\caption{Multi-view 3D reconstruction on ScanNet-50~\cite{scannet}.}
\centering
\scriptsize
\resizebox{0.9\linewidth}{!}{%
\begin{tabular}{c l r ccccc}
\toprule
Frames & \multicolumn{2}{c}{Model} &
Acc.$\downarrow$ & Comp.$\downarrow$ & NC$\uparrow$ &
VRAM(GB)$\downarrow$ & Latency(s)$\downarrow$ \\
\midrule

\multirow{6}{*}[-1.5ex]{300} & \quad VGGT$^*$\cite{VGGT}             &        & \textbf{0.025} & 0.022 & \textbf{0.745} & 18.89 & 62.43 \\
                    & \quad VGGT$^*$\cite{VGGT}                 & + Ours & \textbf{0.025} & \textbf{0.020} & 0.731 & \textbf{12.88} & \textbf{33.70} \\
\cmidrule(lr){2-8}
                    & \quad FasterVGGT\cite{fastervggt}      &        & \textbf{0.029} & 0.025 & \textbf{0.712} & 25.30 & 42.43 \\
                    & \quad FasterVGGT\cite{fastervggt}      & + Ours & 0.034 & \textbf{0.022} & 0.682 & \textbf{13.18} & \textbf{30.08} \\
\cmidrule(lr){2-8}
                    & \quad LiteVGGT\cite{litevggt}          &        & 0.034 & 0.033 & \textbf{0.732} & 16.50 & 17.21 \\
                    & \quad LiteVGGT\cite{litevggt}          & + Ours & \textbf{0.029} & \textbf{0.020} & 0.708 & \textbf{10.04} & \textbf{10.51} \\
\midrule

\multirow{6}{*}[-1.5ex]{500} & \quad VGGT$^*$\cite{VGGT}             &        & 0.029 & 0.024 & \textbf{0.732} & 28.53 & 155.97 \\
                    & \quad VGGT$^*$\cite{VGGT}                 & + Ours & \textbf{0.025} & \textbf{0.019} & 0.724 & \textbf{19.13} & \textbf{59.73} \\
\cmidrule(lr){2-8}
                    & \quad FasterVGGT\cite{fastervggt}      &        & \textbf{0.032} & 0.024 & \textbf{0.703} & 41.02 & 99.87 \\
                    & \quad FasterVGGT\cite{fastervggt}      & + Ours & \textbf{0.032} & \textbf{0.021} & 0.682 & \textbf{15.78} & \textbf{51.94} \\
\cmidrule(lr){2-8}
                    & \quad LiteVGGT\cite{litevggt}          &        & 0.039 & 0.047 & \textbf{0.716} & 26.01 & 36.03 \\
                    & \quad LiteVGGT\cite{litevggt}          & + Ours & \textbf{0.029} & \textbf{0.018} & 0.700 & \textbf{11.98} & \textbf{17.28} \\
\bottomrule
\end{tabular}%
}
\label{tab:3d_full_2}
\end{table}
\section{More Qualitative Results}
\label{sec:supp_qual}

We provide additional qualitative results to complement the quantitative evaluations.
All visualizations are conducted on the NRGBD~\cite{nrgbd} and ScanNet-50~\cite{scannet} datasets.

\subsection{3D Pointmap Reconstructions}
\begin{figure}[t]
\centering
\includegraphics[width=1\linewidth]{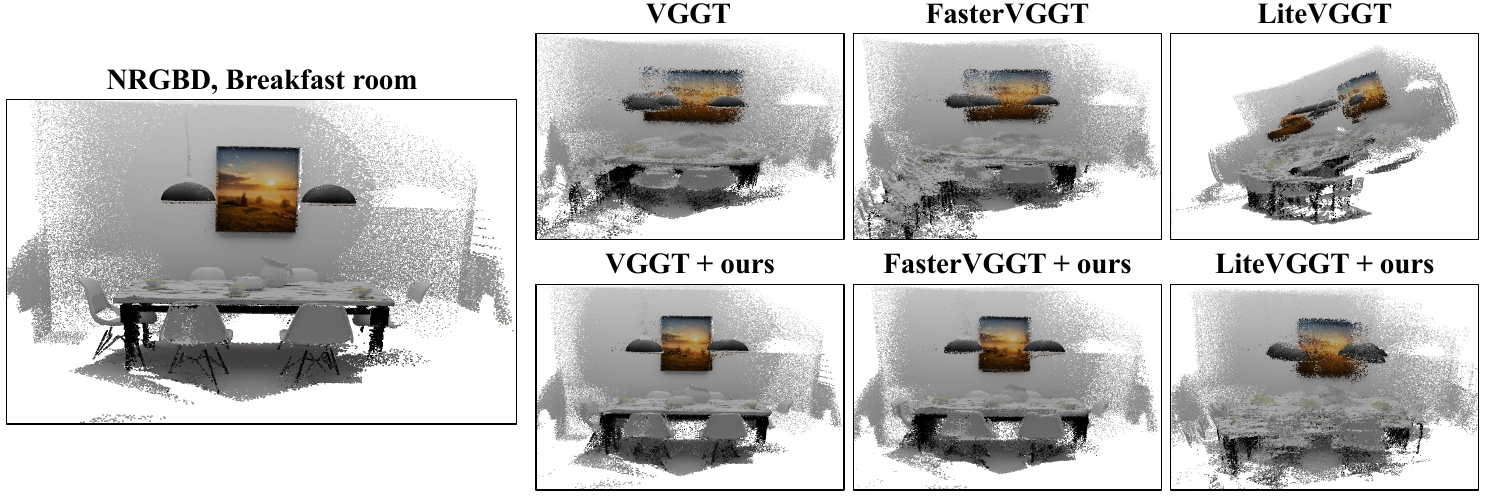}\\[0.3cm]
\includegraphics[width=1\linewidth]{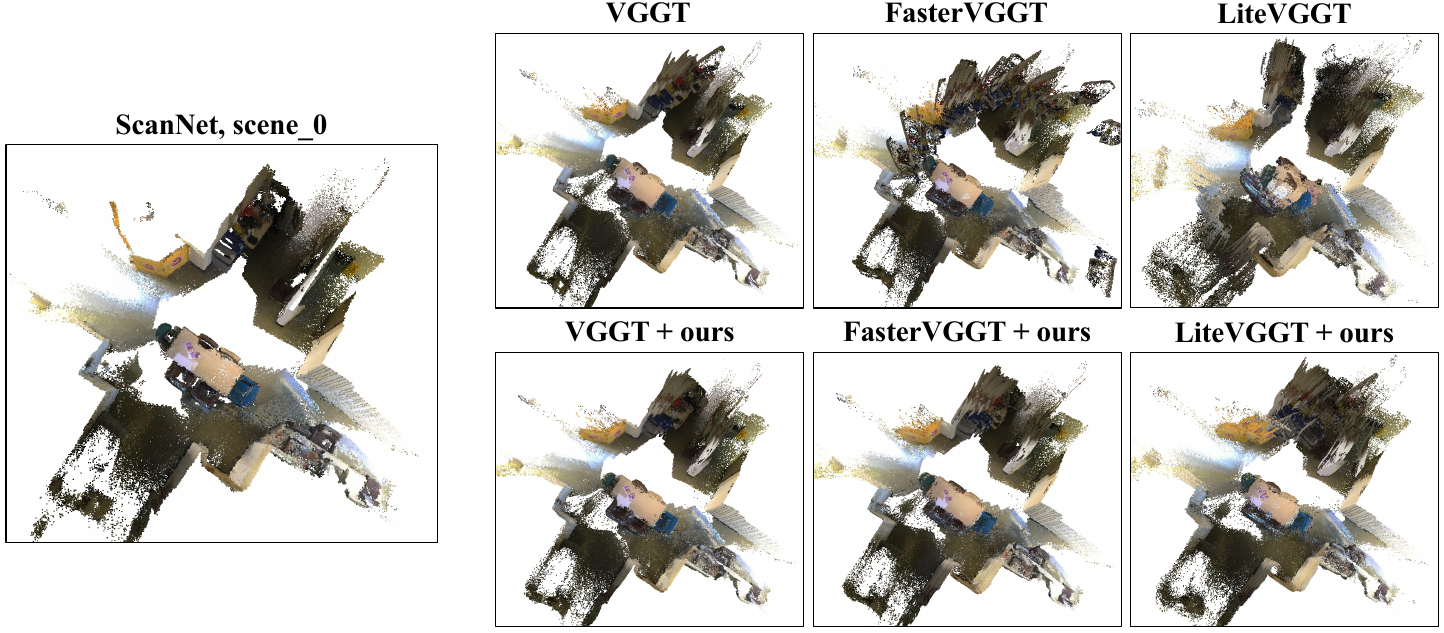}
\caption{3D Point Cloud Reconstruction on NRGBD~\cite{nrgbd} and ScanNet-50~\cite{scannet}.}
\label{fig:supp_fig_structure} 
\end{figure}

\begin{figure}[t]
\centering
\includegraphics[width=1\linewidth]{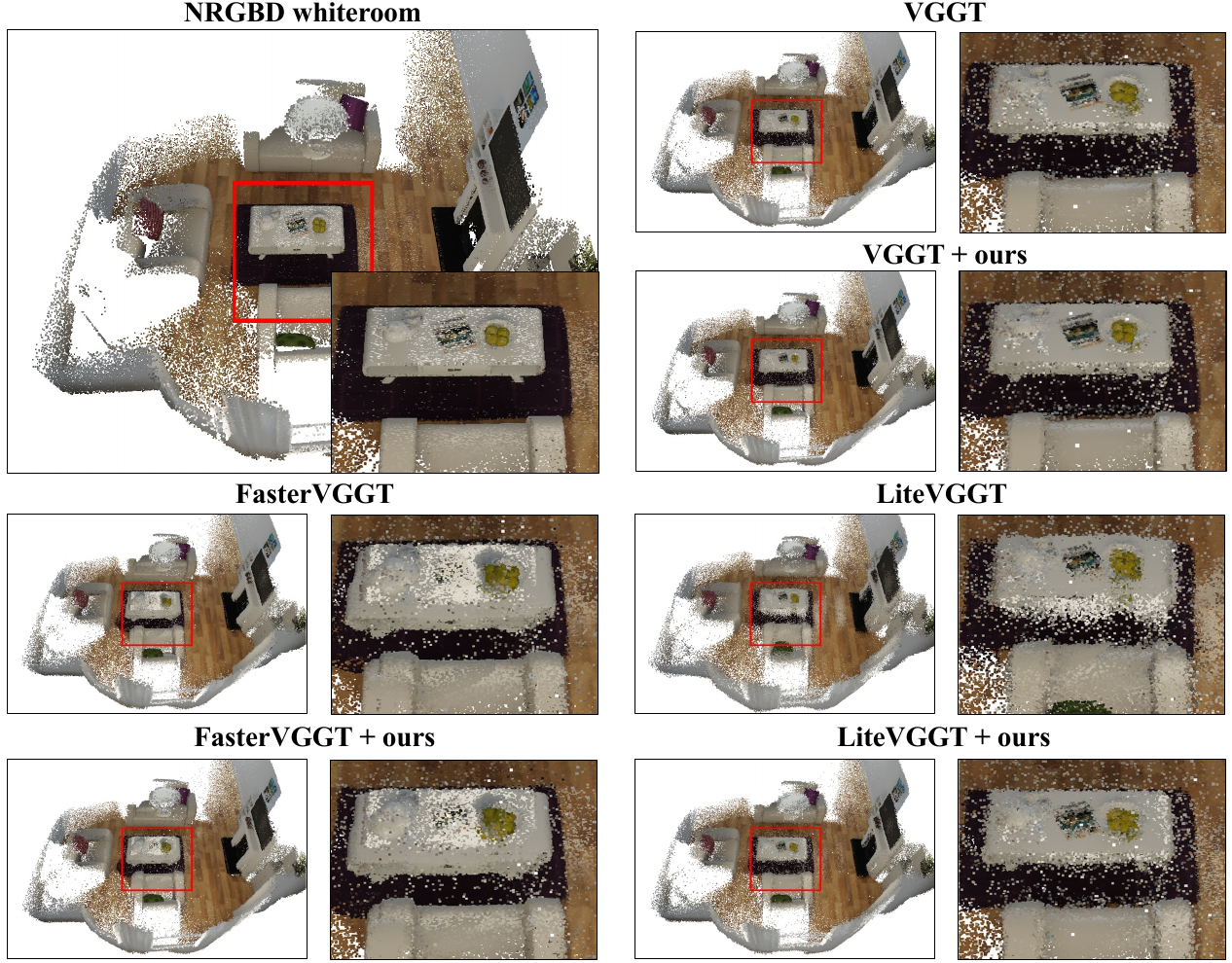}
\caption{Results of 3D Point Cloud Reconstruction on NRGBD~\cite{nrgbd}.}
\label{fig:supp_fig_detail_1} 
\end{figure}

\begin{figure}[t]
\centering
\includegraphics[width=1\linewidth]{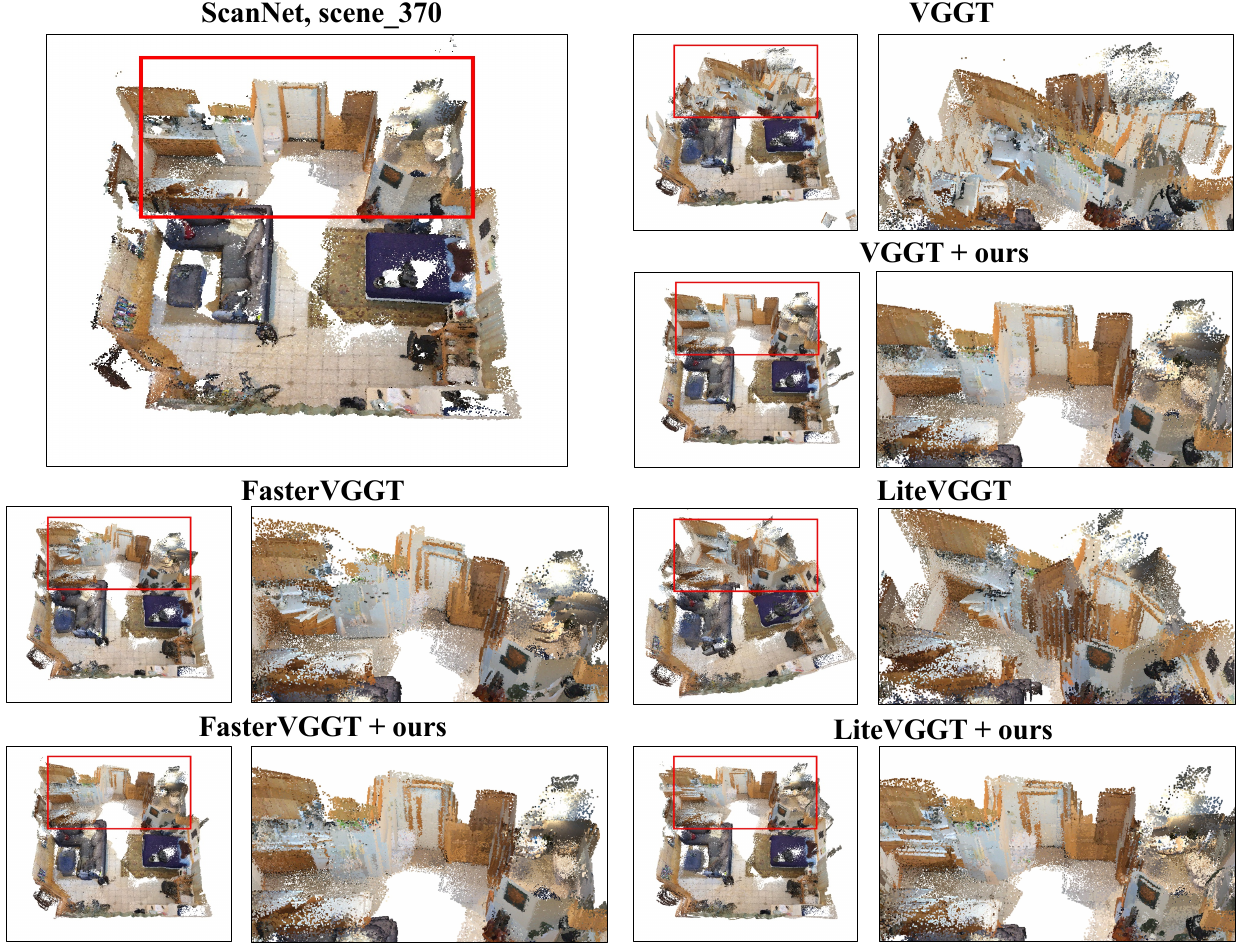}
\caption{Results of 3D Point Cloud Reconstruction on ScanNet-50~\cite{scannet}.}
\label{fig:supp_fig_detail_2} 
\end{figure}

We visualize the reconstructed 3D pointmaps produced by applying our method to VGGT~\cite{VGGT}, FasterVGGT~\cite{fastervggt}, and LiteVGGT~\cite{litevggt}.
As shown in the figures, our method consistently yields more complete and geometrically coherent reconstructions across all backbone models.
Notably, fine-grained structures and scene boundaries are better preserved, demonstrating that our approach effectively enhances reconstruction quality in nearly all cases across different backbone models. As illustrated in Fig.~\ref{fig:supp_fig_structure}, our proposed method yields more structurally coherent reconstructions compared to the baseline models, effectively enhancing global geometric consistency. Furthermore, Fig.~\ref{fig:supp_fig_detail_1} and Fig.~\ref{fig:supp_fig_detail_2} demonstrate that our approach not only preserves but also refines fine-grained geometric details, resulting in higher-fidelity 3D representations.

\subsection{3D Trajectories}
\begin{figure}[t]
\centering
\includegraphics[width=1\linewidth]{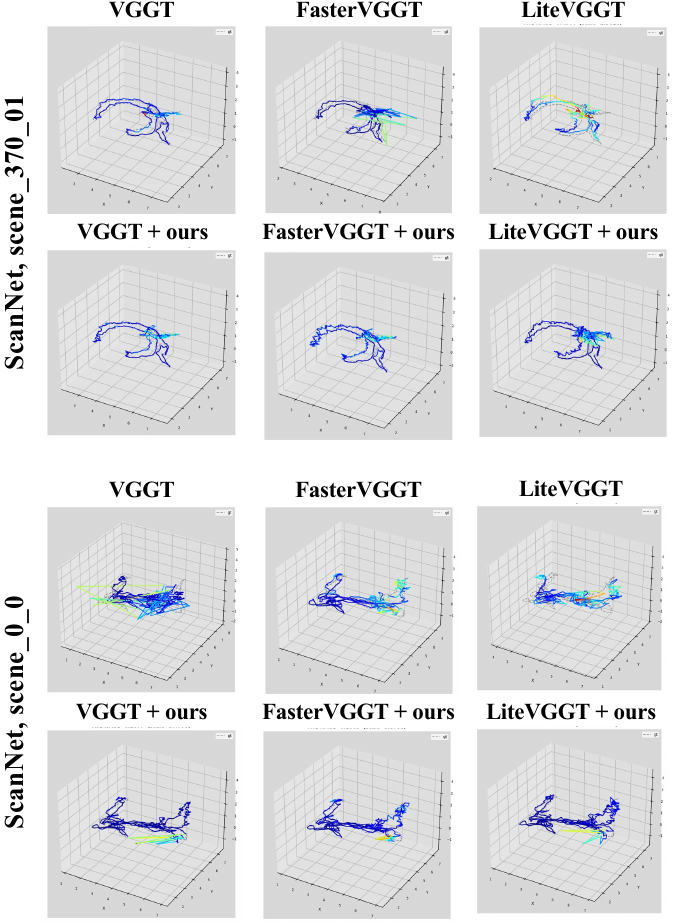}
\caption{Results of 3D Trajectories on ScanNet-50~\cite{scannet}.}
\label{fig:supp_fig_tra} 
\end{figure}

We further compare the estimated camera trajectories with and without our method applied to each backbone.
The resulting trajectories exhibit improved alignment with the ground truth, with reduced drift and fewer outlier poses.
These improvements are consistent across large indoor environments in ScanNet-50~\cite{scannet}.
Qualitative comparisons in Fig.~\ref{fig:supp_fig_tra} demonstrate that our approach yields superior trajectory estimation over the baseline methods. While the baseline models fail to establish accurate camera poses in challenging sequences—leading to completely disordered trajectories—our method robustly estimates the correct poses and maintains global trajectory consistency.

\clearpage

\clearpage
\section{Failure Case}
\label{sec:supp_fail}

\paragraph{\textbf{\textup{Single-Anchor SE(3) Alignment.}}}
Our framework aligns per-chunk reconstructions through SE(3) registration with a single shared anchor chunk (see Sec.~\ref{subsec:all_inference})---a lightweight design that enables scalable, training-free long-sequence inference. While our diversity-aware partitioning naturally yields highly uniform scales across different chunks, minor scale discrepancies can occasionally emerge near partition boundaries since the rigid alignment operates without an explicit global scale optimization.

\begin{figure}[H]
\centering
\includegraphics[width=0.9\linewidth]{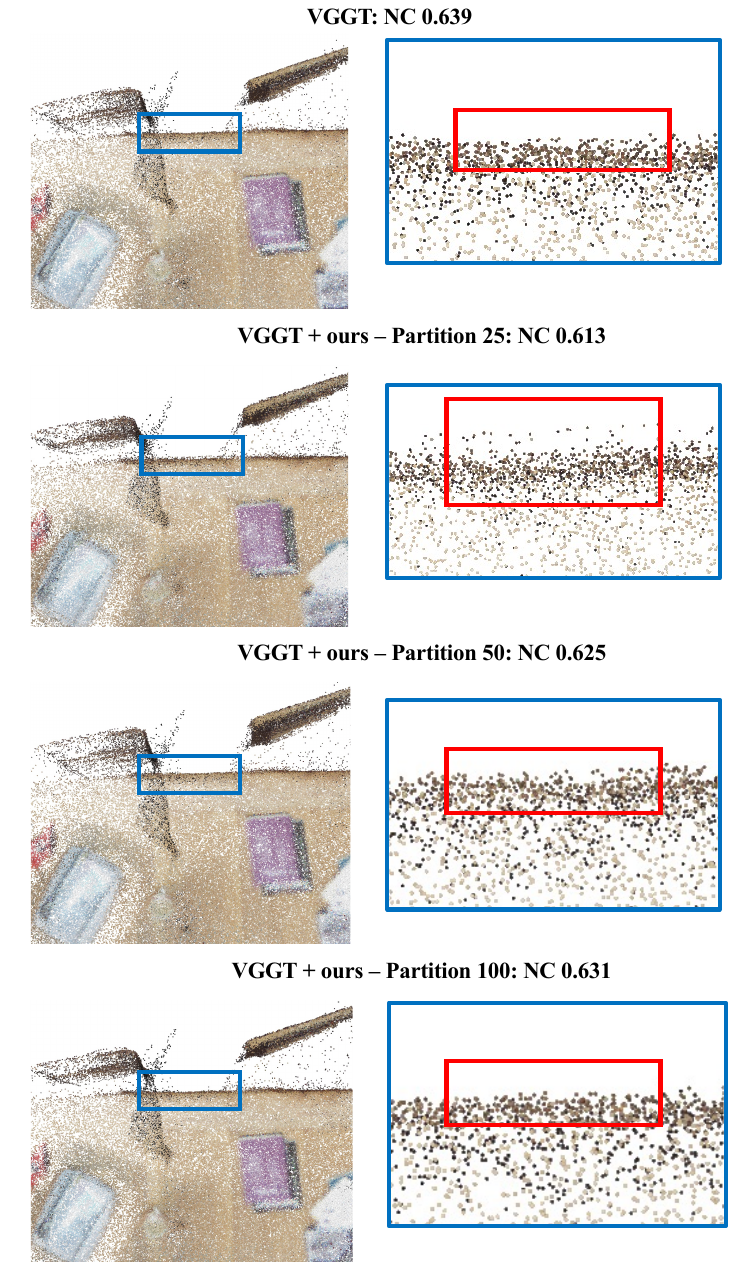}
\caption{Normal Consistency Comparison on 7Scenes~\cite{7scenes}.}
\label{fig:supp_fig_fail} 
\end{figure}

\paragraph{\textbf{\textup{Effect on Metrics.}}}
This slight boundary mismatch is reflected as a marginal drop in \textit{normal consistency} and \textit{multi-view depth}, metrics that are strictly sensitive to fine-grained cross-view alignment. However, this is a trivial artifact rather than a structural defect in the reconstruction itself. As illustrated in Fig.~\ref{fig:supp_fig_fail}, increasing the partition size (e.g., to 100) inherently reduces the number of alignment operations. This naturally alleviates the boundary discrepancies, allowing the normal consistency to seamlessly recover to baseline levels. Furthermore, our method consistently outperforms baselines in camera pose estimation and Chamfer distance, reaffirming that the core geometric reconstruction remains highly accurate.

\paragraph{\textbf{\textup{Discussion.}}}
Ultimately, this slight performance variation is merely a natural and localized consequence of single-anchor alignment, not a fundamental limitation of the model's reconstructive power. While an explicit scale harmonization step could perfectly close this minor gap, our extensive evaluations demonstrate that it is entirely unnecessary for achieving robust and competitive 3D reconstructions.
\newpage

\end{document}